\def\arxivpreprint{}
\documentclass{article}
\usepackage{iclr2026_conference,times}

\usepackage[utf8]{inputenc}
\usepackage[T1]{fontenc}
\usepackage{algorithm}
\usepackage{algpseudocode}
\usepackage{xspace}
\usepackage{booktabs}
\usepackage{multirow}
\usepackage{tcolorbox}
\usepackage{wrapfig}
\usepackage{graphicx}
\usepackage{float}
\usepackage{amsmath}
\usepackage{amsfonts}
\usepackage{amsthm}
\usepackage{enumitem}
\usepackage{array}
\usepackage{tikz}
\usepackage{subcaption}
\usepackage{nicefrac}
\usepackage{microtype}
\usepackage{xcolor}
\usepackage{hyperref}
\usepackage{url}

\newtheorem{definition}{Definition}

\newcommand{\method}{\textsc{Pearl}\xspace}
\newcommand{\dataset}{\textsc{PMH-IG}\xspace}

\title{Personalized Image Generation with Reasoning and Reflection}
\hypersetup{pdftitle={Personalized Image Generation with Reasoning and Reflection},
            hidelinks}
\ifdefined\arxivpreprint
  \iclrfinalcopy
  \title{Personalized Image Generation\\with Reasoning and Reflection\par\vspace{10pt}}
  \author{%
  \parbox{\dimexpr\textwidth-2\tabcolsep\relax}{%
    \centering
    Bo Ni\textsuperscript{1},
    Ngoc N. Tran\textsuperscript{1},
    Qinwen Ge\textsuperscript{1},
    Franck Dernoncourt\textsuperscript{2}\\[3pt]
    Seunghyun Yoon\textsuperscript{2},
    Samyadeep Basu\textsuperscript{3},
    Sungchul Kim\textsuperscript{2},
    Puneet Mathur\textsuperscript{2}\\[3pt]
    Nedim Lipka\textsuperscript{2},
    Tong Yu\textsuperscript{2},
    Yu Wang\textsuperscript{4},
    Ryan A. Rossi\textsuperscript{2},
    Tyler Derr\textsuperscript{1}\\[8pt]
    \normalfont
    \textsuperscript{1}Vanderbilt University\quad
    \textsuperscript{2}Adobe Systems\\[2pt]
    \textsuperscript{3}University of Maryland, College Park\\[2pt]
    \textsuperscript{4}University of Georgia
  }%
}
\hypersetup{pdfauthor={Bo Ni, Ngoc N. Tran, Qinwen Ge, Franck Dernoncourt,
  Seunghyun Yoon, Samyadeep Basu, Sungchul Kim, Puneet Mathur, Nedim Lipka,
  Tong Yu, Yu Wang, Ryan A. Rossi, Tyler Derr}}

\else
  \author{Anonymous authors}
  \hypersetup{pdfauthor={}}
\fi

\begin{document}
\maketitle
\ifdefined\arxivpreprint
  \lhead{Preprint}
\fi
\begin{abstract}
Personalized image generation has remained narrowly focused on conditional synthesis from curated visual exemplars, rather than capturing who a user is. In practice, however, a user's personal context is much richer, comprising reviews, posts, images, captions, and metadata accumulated over time. A truly personalized generator should leverage this history to produce images aligned with the user's lifestyle and aesthetic preferences. To this end, we introduce the first unified benchmark for personalized image generation from user histories. The benchmark comprises two complementary tasks and a multi-axis evaluation protocol that assesses target fidelity, visual quality, user distinguishability, semantic alignment with the user's history, and task-specific utility. Grounded in real-world e-commerce and social media settings, the benchmark includes: (1) \emph{Personalized Scene Generation}, which places a given object in a scene that reflects a user's preferences and lifestyle, motivated by personalized product presentation; and (2) \emph{Personalized Creative Generation}, which generates a novel image on a specified topic that is faithful to a user's aesthetic and visual identity, motivated by social media content creation. We further propose \method , which couples a multimodal reasoner with a frozen image generator in an interleaved \emph{reason--reflect} loop optimized with differential data reward. Across both tasks, \method
outperforms strong baselines, achieving an average improvement of 15\% across personalization metrics.
\end{abstract}

\section{Introduction}
In modern AI systems, adapting outputs to individual users' needs, preferences, and communication styles is essential for producing content that is relevant, engaging, and broadly useful~\citep{salemi2024lamplargelanguagemodels, zhang2025personalizationlargelanguagemodels}. This principle of user modeling has long been studied in the information retrieval~\citep{10.1145/1076034.1076111, 10.1145/2348283.2348312}, human-computer interaction~\citep{SCHIAFFINO2004129}, and recommender~\citep{naumov2019deep, he2017neural, koren2009matrix} communities, where systems such as search engines, news feeds, and product recommenders rely on user histories to deliver personalized experiences. More recently, the same principle has been extended to large language models through personalized text generation, where conditioning generation on a user's historical posts, reviews, and interactions substantially improves the relevance and quality of the output~\citep{ni2026reasoningbasedpersonalizedgenerationusers, xu2025personalized}.

Building on this line of work, recent personalized text generation methods~\citep{salemi2024lamplargelanguagemodels, au2025personalizedgraphbasedretrievallarge,
ni2026reasoningbasedpersonalizedgenerationusers, li2024personalizedlanguagemodelingpersonalized, kumar2024longlampbenchmarkpersonalizedlongform} leverage a user's long history to produce text that mirrors that user's writing style, topical interests, and stated preferences. In contrast, personalized image generation has remained focused on conditioned image generation~\citep{wei2025personalizedimagegenerationdeep,
ruiz2023dreamboothfinetuningtexttoimage, yang2023controllabletextualinversionpersonalized, ye2023ipadaptertextcompatibleimage}, where \textit{personalization} is
defined as reproducing specific visual concepts (e.g., faces or painting styles)
from a small set of reference images while adhering to novel text prompts. However, real-world personalization requires a richer understanding of \textit{who the user is} to generate images that align with the user's lifestyle, interests, and aesthetic preferences. For example, in e-commerce, a personalized product background generated based on the user's inferred preferences can drive higher click-through rates than generic catalog images~\citep{Czapp_2024, 10.1145/3240323.3241729}. Similarly, social media users cultivate distinctive visual identities through their posts, so generated content that aligns with this identity is more likely to integrate naturally into their feeds and attract more engagement~\citep{moller2026impact}. 

To bridge this gap, we introduce \dataset(\textbf{P}ersonalized \textbf{M}ulti-modal \textbf{H}istory conditioned \textbf{I}mage \textbf{G}eneration), a unified benchmark for personalized image generation from realistic user histories. \dataset{} consists of in-the-wild rich user histories collected from online reviews and social media that can be utilized for identity inference and downstream image generation. Compared to existing benchmarks~\citep{xu2025personalized, dunlop2025personalized}, which provide a few curated reference images per user and evaluate concept reproduction under novel prompts, \dataset{} provides realistic user histories and evaluates whether a model can infer the user's identity and synthesize a new image that reflects it. \dataset{} instantiates identity-aware image generation through two complementary tasks: \emph{Personalized Scene Generation} and \emph{Personalized Creative Generation}, where the model must render a target product or topic into an image grounded in the user's history. The two tasks cover the principal axes of real-world identity: \emph{Personalized Scene Generation} draws on Amazon review histories, which encode \emph{what} a user does through textual behavioral signals, while \emph{Personalized Creative Generation} draws on Instagram post histories, which encode \emph{how} a user looks through visual aesthetic signals.
Our experiments show that current personalization methods struggle on both tasks, as they cannot reason over the complex user history and translate it into an appropriate image.

We thus further propose \method, a reasoning-interleaved framework that addresses the core limitation underlying these failures. Current personalization frameworks~\citep{xu2025personalized, shen2024pmgpersonalizedmultimodal} treat the user history as a unified conditioning signal and fuse it directly into the generator, bypassing any explicit reasoning about what the history implies for the scene, the user's identity, or the aesthetic that should govern the output. This leaves personalization cues entangled in dense vectors that the generator must decode implicitly. An alternative is to externalize this inference with a multimodal reasoner that consumes the history and emits a textual image-generation plan for a frozen text-to-image generator to render, but the plan is written without knowing how the generator will instantiate it, so personalization cues correctly identified in the history can still be lost at render time. \method{} addresses the challenge by first reasoning over the user's history to produce an explicit image generation plan and an initial rendering, then reflecting on that rendering by comparing it against the history to identify concrete personalization mismatches, and finally re-renders from a corrected plan that is grounded in the first rendering. \method{} is trained through a two-stage cross-modal reflection tuning procedure that first learns to plan from user history and then to revise plans in light of rendered evidence. In summary, our contribution can be summarized as follows:

\begin{figure}[t]
    \centering
    \includegraphics[width=0.9\linewidth]{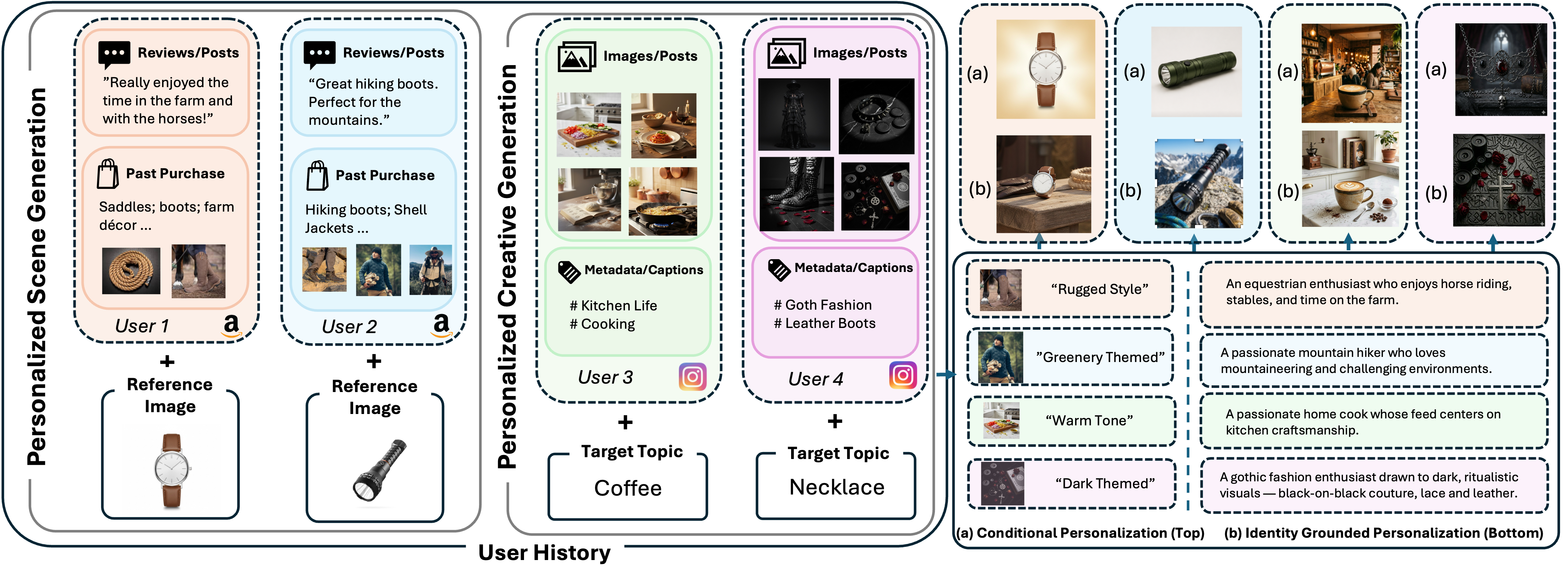}
    \vspace{-1ex}
    \caption{Personalized Image Generation based on User History.}
    \label{fig:placeholder}
    \vskip -3.5ex
\end{figure}

\vspace{-1.5ex}
\begin{itemize}[leftmargin=*]
    \item We introduce \dataset, the first benchmark for personalized image generation from realistic user histories, which comprises two complementary tasks, \emph{Personalized Scene Generation} on Amazon review histories and \emph{Personalized Creative Generation} on Instagram post histories.

    \vspace{-0.5ex}
    \item We propose \method, a novel reasoning-interleaved framework for identity-aware image generation that first reasons over a user's history to produce an explicit image-generation plan and an initial rendering, then reflects on that rendering against the history to identify personalization mismatches, and finally re-renders from a corrected plan. The reasoner is trained through a two-stage cross-modal reflection tuning procedure.

    \vspace{-0.5ex}
    \item We conduct extensive experiments along three evaluation axes: Image generation quality, personalization, and MLLM-as-a-Judge that directly probes lifestyle alignment. \method{} substantially outperforms strong personalization baselines on both tasks, and ablations confirm that both reasoning and reflection contribute measurable gains.
\end{itemize}

The rest of the paper will be organized as follows: Section~\ref{sec:dataset} formally introduces the proposed problem definition and benchmark details. Section~\ref{sec:method} outlines the proposed \method framework, and Section~\ref{sec:experiments} reports the extensive experiment results and analysis. Section~\ref{sec:related-work} further positions our work within the existing literature and we conclude in Section~\ref{sec:conclusion}. 

\section{A Benchmark for Personalized Image Generation from User Histories}
\label{sec:dataset}

In this section, we introduce \dataset, a new benchmark for personalized image generation from multimodal user histories. Building from public user activities on online platforms, \dataset{} includes two complementary tasks.
\textit{Personalized Scene Generation} asks a model to render a given object in a scene that reflects a user's inferred lifestyle and preferences, supporting applications such as personalized product presentation and generative recommendation.
\textit{Personalized Creative Generation} asks a model to generate a new image conditioned on a topic or content specification while preserving the user's established visual identity, supporting applications such as personalized content creation and creator tooling.
In the rest of this section, we will first formally define the general problem setting, then describe the benchmark design principles and the two task instantiations. 
Lastly, we introduce an overview of the evaluation protocol; detailed metric definitions
are provided in Section~\ref{sec:experiments}. 
The more detailed descriptive statistics for the datasets are provided in Appendix~\ref{app:dataset}. 

\subsection{Problem Definition}
\label{sec:problem}

We study personalized image generation grounded in a user's accumulated history. Different from prior studies on conditional concept personalization~\citep{shen2024pmgpersonalizedmultimodal, xu2025personalized, wei2025personalizedimagegenerationdeep}, where the reference typically depicts a single visual concept (a face, object, or painting style), we focus on identity inference from a user's naturally occurring multimodal history, where the model must aggregate distributed behavioral and aesthetic cues across prior activities
rather than reproduce any single depicted concept.

For each user $u$, let $\mathcal{H}_u = \{h^{(u)}_1, h^{(u)}_2, \ldots, h^{(u)}_{N_u}\}$ be the user's history, where each entry $h^{(u)}_i$ records one prior activity,
such as a written review with an associated product image or a social media post with its caption. We consider a personalized generation task $T$, which defines the form of personalized image generation to perform and has an associated condition space $\mathcal{C}_T$. For a given task $T$, the model receives the user history $\mathcal{H}_u$ and a task-specific generation condition $c\in\mathcal{C}_T$, where $c$ specifies what the generated image should depict or preserve. The full notation is summarized in Appendix~\ref{app:notation}.

\begin{definition}[Personalized Image Generation from User History]
Given a personalized generation task $T$ with condition space $\mathcal{C}_T$, the objective is to learn a task-specific parameterized model $f^T_\theta$ that maps a user's multimodal history $\mathcal{H}_u$ and a generation condition $c\in\mathcal{C}_T$ to a personalized image
\begin{equation}
\hat{I}_{u,c}=f^T_\theta(\mathcal{H}_u,c),
\label{eq:problem}
\end{equation}
such that $\hat{I}_{u,c}$ follows the generation condition $c$ and reflects the user's preferences, behaviors, and aesthetics expressed in $\mathcal{H}_u$.
\end{definition}

We instantiate this general problem through two complementary tasks that probe different axes of user identity, with task-specific datasets described in Sections~\ref{sec:scene} and~\ref{sec:creative} (with additional details, such as data curation and instance construction are provided in Appendix~\ref{app:dataset}). 

\begin{itemize}[leftmargin=*]
\item \noindent \textbf{Task: Personalized Scene Generation.}
The generation condition is a product or object image, and the model $f^T_\theta$ must preserve the object while placing it in a scene coherent with the user's inferred lifestyle and preferences.
This task probes the model's ability to identify a user's recurring activities, interests, and lifestyle contexts from the behavioral signals in $\mathcal{H}_u$ and translate them into a coherent visual scene around the target product.

\vspace{-0.5ex}
\item \noindent \textbf{Task: Personalized Creative Generation.}
The generation condition is a topic prompt or content specification, and the model $f^T_\theta$ must depict the requested topic while matching the user's established visual identity.
This task probes the model's ability to capture a user's aesthetic identity from past visual posts and metadata in $\mathcal{H}_u$, apply it to novel topics, and generate content that fits within the user's visual history.

\end{itemize}

\vspace{-1.5ex}
\subsection{Benchmark Design Principles}
\vspace{-1ex}
\label{sec:principles}

To operationalize the problem definition,
we construct 
\dataset based on three design principles, and each corresponds to a key benchmark choice: \emph{where} personalization signals come from, \emph{what} role the generation condition plays, and \emph{how} generated images are evaluated.
Together they ensure that \dataset rewards genuine identity inference with balanced image quality and personalization.

\begin{itemize}[leftmargin=*]
\item \textbf{Naturally occurring user histories.} First, the personalization signal should come from a user's naturally occurring activity history rather than curated exemplars or synthetic preference templates. Real personalization requires reasoning over the noisy, multimodal, and behaviorally grounded traces that users leave on platforms, which 
is the very personalization cue we aim to study. 
\dataset{} is
therefore built exclusively on public user data, such as Amazon reviews and Instagram posts, 
with each history entry corresponding to a real review or post authored by a real
account. 

\item \textbf{User signal--generation condition distinction.} 
Second, generation condition $c$ and user history $\mathcal{H}_u$ play distinct roles. The condition specifies what should be generated, while the user history provides the personalization signal that determines how it should be generated. In Personalized Scene Generation, the product image determines the object to render, while the user history determines the scene context. In Personalized Creative Image Generation, the topic determines the content, while the user history determines the visual identity. Thus, the user plays the role of context rather than subject, separating \dataset from concept personalization, where conditioning images depict the subject to reproduce under new prompts.

\item \noindent \textbf{Multi-axis evaluation.} 
Third, evaluation must jointly measure condition fidelity, visual quality, and user-level personalization, since each can be satisfied without the others. A clean image that follows the generation condition may still carry no user-specific signal, while an image that reflects the user may fail to preserve personalization or maintain visual fidelity. \dataset therefore pairs image-based measures with personalization and task-specific utility measures, including retrieval-based and judge-based evaluations. We introduce these axes in Section~\ref{sec:eval} and Appendix~\ref{app:evaluation}.

\end{itemize}

\vspace{-0.5ex}
\subsection{Personalized Scene Generation: E-commerce Product Presentation}
\vspace{-0.5ex}
\label{sec:scene}
We instantiate Personalized Scene Generation in the e-commerce setting, where a product is presented to a user through a generated lifestyle scene. The user history $\mathcal{H}_u=\{(\ell_i,o_i)\}_{i=1}^{n}$ is drawn from Amazon reviews, with each entry pairing a written review $\ell_i$ with the corresponding reviewed product image $o_i$. The generation condition $c=o^\star$ is the product image to be personalized, and the model must generate a scene that preserves the product while adapting its surrounding context to the user's inferred interests and lifestyle. The same product should yield visibly different scenes for different users: a flashlight should appear on a hiking trail for a user whose history is dominated by camping gear, but perhaps in an engine bay for a user whose activities concentrate on home automotive repair. 

\vspace{-0.5ex}
\subsection{Personalized Creative Generation: Social Media Content Posting}
\label{sec:creative}
\vspace{-0.5ex}
We instantiate Personalized Creative Generation in a social media setting, where a model generates new visual content that fits a user's established posting style. The user history $\mathcal{H}_u=\{(z_i,m_i)\}_{i=1}^{n}$ is drawn from Instagram posts, with each entry pairing a social media image $z_i$ with associated caption metadata $m_i$. The generation condition $c=t^\star$ is a topic or content specification, and the model must generate an image that depicts the requested topic while matching the user's recurring visual identity. The same topic should yield visually different images for different users, reflecting differences in composition, subject framing, color, setting, and presentation style. 

\vspace{-0.5ex}
\subsection{Evaluation Protocol}
\label{sec:eval}
\vspace{-0.5ex}

\noindent \textbf{Shared Metrics.}
We evaluate generations along two axes shared across both tasks: \emph{output quality}, captured by the LAION aesthetic predictor~\citep{schuhmann2022laion5bopenlargescaledataset} as a reference-free image quality score, and \emph{holistic user alignment}, captured by an MLLM-as-Judge protocol that scores each generation along style, content, and overall dimensions on a $1$--$5$ scale. To mitigate single-model bias, the MLLM judge is an ensemble of Gemini~Flash~2.5~\citep{comanici2025gemini25pushingfrontier} and Qwen2.5-VL-32B~\citep{bai2023qwenvlversatilevisionlanguagemodel}; we report the average of the two models' scores. Both shared metrics are reference-free, so they apply uniformly to the two tasks.

\noindent \textbf{Personalized Scene Generation Specific Metrics.}
Because Personalized Scene Generation has no naturally occurring ground-truth scene image per (user, product) pair, we evaluate user alignment through a contrastive recommendation lens. We train a two-tower contrastive visual recommender on real (review history, purchased product image) pairs from the training split, encode each generation with the image tower, and report Hit@$5$ and MRR over a retrieval pool containing the user's held-out true purchase. The metrics ask whether a generation is sufficient on its own to recommend the right product back to its user. 

\vspace{-0.5ex}
\paragraph{Personalized Creative Generation Specific Metrics.}
Personalized Creative Generation has a natural ground-truth target (the user's actual held-out post), enabling reference-based metrics in addition to user alignment. For fidelity to target, we report CLIP image similarity (CIS), DINO image similarity (DIS), LPIPS, and MS-SSIM between each generation and the held-out post. For user alignment, we train a StyleDiscriminator on (history, next post) pairs from training users and ask it to retrieve the correct user from a decoy pool, stratified at two difficulty levels: \emph{inter-category} R@$1$, with decoys drawn from users in different topical categories, and \emph{intra-category} R@$1$, with decoys sharing the topical category but differing in visual style.

Full metric definitions, retrieval-pool construction and MLLM-judge prompts are in Appendix~\ref{app:evaluation}.
\vspace{-1ex}
\section{\method: Personalized Image Generation with Reasoning and Reflection}
\label{sec:method}
\begin{figure}
      \centering
      \includegraphics[width=\linewidth]{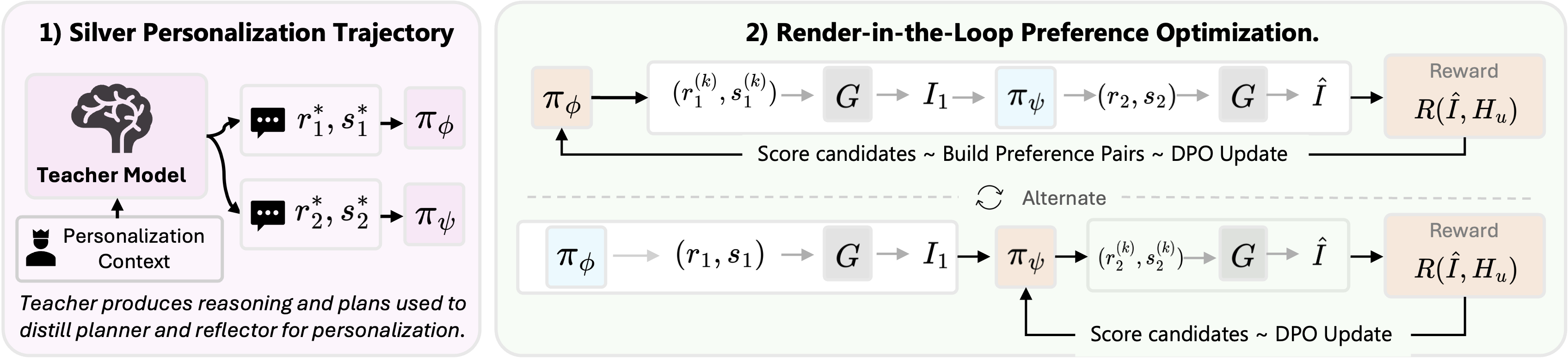}
      \caption{Training procedure of the \method framework. We warm start with the Silver Personalization Trajectory, and optimize both policies with the Render-in-the-Loop optimization.}
      \label{fig:pearl-training}
  \end{figure}

\vspace{-1ex}
We introduce \method, a reasoning-interleaved framework for personalized image generation (Figure~\ref{fig:pearl-training}). A Stage-1 \emph{planner} reasons over user history to produce a scene plan, which a frozen renderer turns into an image. A Stage-2 \emph{reflector} compares this image with the history, identifies personalization mismatches, and revises the plan for the same renderer, grounding personalization in rendered evidence. Both policies are jointly trained via alternating-policy DPO with a task-aligned retrieval reward, each optimized against the renderer and the other policy used at inference. Sections~\ref{sec:method-planner}--\ref{sec:method-reflector} detail the policies, Section~\ref{sec:method-train} describes training, and Appendix~\ref{app:algorithms} provides pseudocode.

\vspace{-0.5ex}
\subsection{Planner: Reasoning over User History}
\label{sec:method-planner}
\vspace{-0.5ex}
Given the user history $\mathcal{H}_u$, the planner translates it into a
concrete specification that an image generator can execute. Since the
relevant scene for a generation condition $c$ is rarely depicted in any single entry
of $\mathcal{H}_u$, this requires reasoning over the full history rather
than retrieval from it: the planner must aggregate behavioral and
aesthetic signals scattered across many entries and commit them to a
single coherent scene description. We instantiate the planner as a
multimodal policy $\pi_{\phi}$ that, given $(\mathcal{H}_u, c)$,
autoregressively emits an interleaved output consisting of a
chain-of-thought reasoning trace $r_1$ followed by a scene description
$s_1$, 
\begin{equation}
    (r_1, s_1) = \pi_{\phi}(\cdot \mid \mathcal{H}_u, c),
\end{equation}
where $r_1$ summarizes what the history implies for the user's lifestyle
and aesthetic, and $s_1$ is a self-contained text prompt suitable as
input to a text-to-image generator. A frozen image generator $G$ then
produces the initial rendering $I_1 = G(s_1, c)$.

\vspace{-0.5ex}
\subsection{Reflector: Cross-Modal Refinement from Rendered Evidence}
\label{sec:method-reflector}
\vspace{-0.5ex}
The initial rendering $I_1$ generated from the scene description $s_1$
can deviate from the user's identity in ways the planner could not
anticipate, because $\pi_\phi$ commits to $s_1$ before observing how
$G$ instantiates it. Personalization cues from $\mathcal{H}_u$ can thus
be lost or distorted at render time. To further align \method{} with the user's
identity, we introduce a reflector that conditions on the rendered
image $I_1$ as additional evidence and revises the scene description
accordingly. Let $\pi_\psi$ be the reflector. Given $(\mathcal{H}_u,
I_1)$, $\pi_\psi$ autoregressively emits an interleaved output
$(r_2, s_2) = \pi_\psi(\cdot \mid \mathcal{H}_u, I_1)$ consisting of a structured
analysis $r_2$ followed by a revised scene description $s_2$, where
$r_2$ compares $I_1$ against $\mathcal{H}_u$ along concrete
personalization axes, identifying mismatches such as scene
incongruities, missing lifestyle cues, or aesthetic deviations from the
user's established style, and $s_2$ is a self-contained text prompt
formatted identically to $s_1$. We then generate the final image
$\hat{I}$,
\begin{equation}
    \hat{I} = G(s_2, c)
    \label{eq:reflector-render}
\end{equation}

\subsection{Training}
\label{sec:method-train}
We train \method{} in two stages. In the first stage, we initialize $\pi_\phi$ and $\pi_\psi$ via supervised fine-tuning on \textit{silver} personalization trajectories distilled from a teacher multimodal model with access to the target image. In the second stage, we jointly refine both policies via \emph{render-in-the-loop} preference optimization, where each policy is scored by the full downstream pipeline that includes the other policy. We describe each stage in detail below. The training pseudocode is provided in Appendix~\ref{app:algorithms}.

\paragraph{Stage 1: Silver Personalization Trajectory Distillation.}
We construct silver supervision for both policies by prompting a teacher multi-modal model with a ground-truth identity signal $\mathcal{I}_u^\star$ alongside the user history $\mathcal{H}_u$. Concretely, $\mathcal{I}_u^\star$ may be a held-out image consistent with the user's identity or a structured set of identity elements extracted from $\mathcal{H}_u$, depending on the task specification. For the planner, the teacher receives $(\mathcal{H}_u, c, \mathcal{I}_u^\star)$ and produces a silver trajectory $(r_1^\star, s_1^\star)$ that a student conditioned on $(\mathcal{H}_u, c)$ alone could plausibly reproduce. For the reflector, we render $s_1^\star$ through $G$ to obtain a pseudo-initial image $\tilde{I}_1$, then prompt the teacher with $(\mathcal{H}_u, \tilde{I}_1, \mathcal{I}_u^\star)$ to produce a silver trajectory $(r_2^\star, s_2^\star)$ describing how $\tilde{I}_1$ should be corrected. The two policies are then independently fine-tuned on their respective silver trajectories via standard cross-entropy:
{\small   
\begin{equation*}
    \mathcal{L}_{\text{SFT}}(\pi_\phi) = -\mathbb{E}_{(\mathcal{H}_u, c)}\!\left[\log \pi_\phi(r_1^\star, s_1^\star \mid \mathcal{H}_u, c)\right], \quad 
    \mathcal{L}_{\text{SFT}}(\pi_\psi) = -\mathbb{E}_{(\mathcal{H}_u, \tilde{I}_1)}\!\left[\log \pi_\psi(r_2^\star, s_2^\star \mid \mathcal{H}_u, \tilde{I}_1, c)\right].
\end{equation*}}

\paragraph{Stage 2: Render-in-the-Loop Preference Optimization.}
The supervised warm start trains each policy independently, but at inference the two policies compose: $\pi_\phi$'s output is consumed by $\pi_\psi$ and $G$, and $\pi_\psi$'s output depends on what $\pi_\phi$ produced. We address this by alternating between updating $\pi_\phi$ and $\pi_\psi$ via Direct Preference Optimization~\citep{rafailov2024directpreferenceoptimizationlanguage}, with the other policy and the renderer held frozen at each step, and with each candidate scored by the actual downstream generation. Concretely, for the planner update, we sample $K$ candidate planner outputs $\{(r_1^{(k)}, s_1^{(k)})\}_{k=1}^K \sim \pi_\phi(\cdot \mid \mathcal{H}_u, c)$, run the full downstream pipeline $\pi_\psi \circ G$ on each to obtain a final image $\hat{I}^{(k)}$, and score it with a task-aligned retrieval reward $R(\hat{I}^{(k)}, \mathcal{H}_u)$ (defined in Section~\ref{sec:experiments}) that measures how well the image fits the user. We form preference pairs from the highest- and lowest-scoring candidates, $(r_1^{w}, s_1^{w}) \succ (r_1^{l}, s_1^{l})$, and update $\pi_\phi$ with the standard DPO objective:
\begin{equation}
\label{eq:dpo}
    \mathcal{L}_{\text{DPO}}(\pi_\phi) = -\mathbb{E}\!\left[\log \sigma\!\left(\beta \log \frac{\pi_\phi(r_1^{w}, s_1^{w} \mid \mathcal{H}_u, c)}{\pi_\phi^{\text{ref}}(r_1^{w}, s_1^{w} \mid \mathcal{H}_u, c)} - \beta \log \frac{\pi_\phi(r_1^{l}, s_1^{l} \mid \mathcal{H}_u, c)}{\pi_\phi^{\text{ref}}(r_1^{l}, s_1^{l} \mid \mathcal{H}_u, c)}\right)\right],
\end{equation}
where $\pi_\phi^{\text{ref}}$ is the warm-started checkpoint and $\beta$ is the DPO temperature. The reflector update reuses
Equation~\ref{eq:dpo} with $\pi_\phi$ taken to be $\pi_\psi$ and the
conditioning context $(\mathcal{H}_u, c)$ taken to be
$(\mathcal{H}_u, I_1)$, where $I_1 = G(s_1, c)$ is the initial
rendering produced by the frozen planner; preference pairs are formed
by sampling $K$ candidate continuations from $\pi_\psi(\cdot \mid
\mathcal{H}_u, I_1)$, rendering each $s_2$ through $G$ to obtain
$\hat{I}$, and ranking under $R$.

We perform one planner update followed by one reflector update; while alternating for additional rounds is straightforward and we expect it to yield further gains, we leave a systematic study of multi-round optimization to future work. Even with a single round, because each policy is scored by the full downstream pipeline that includes the other policy, every update step optimizes against the exact composition encountered at inference rather than against a fixed teacher signal.
\section{Experiments}
\label{sec:experiments}
\vspace{-0.5ex}
We conduct extensive experiments to investigate how existing personalization methods perform on \dataset and the effectiveness of the proposed \method. We will first introduce the baseline and experiment set-up. Then we will report the experiment results and analysis. 

\vspace{-0.5ex}
\subsection{Baselines}
\vspace{-0.5ex}
\label{sec:baselines}
We compare against two prior history-based personalization methods and two pure Multimodal Large Language Models (MLLM) baselines.

\vspace{-0.5ex}
\begin{itemize}[leftmargin=*]

\vspace{-0.5ex}
\item  \textbf{PMG}~\citep{shen2024pmgpersonalizedmultimodal}: SDXL conditioned on LLM-extracted style keywords from user history.

\item \textbf{Pigeon}~\citep{xu2025personalized}: a recent MLLM-based personalization baseline that applies a learned masking on the user history before conditioning.

\vspace{-0.5ex}
\item \textbf{LLaVA}~\citep{liu2023visualinstructiontuning}: an MLLM designed to extract dense image features for visual reasoning, generating text by default but
producing images with an external text-to-image generator. 

\vspace{-0.5ex}
\item \textbf{LaVIT}~\citep{jin2024unifiedlanguagevisionpretrainingllm}: an MLLM that converts images into discrete visual tokens for reasoning and generates visual tokens to guide the
image generation.
We use it out-of-the-box with no additional training; visual tokens and a category string are decoded by its native SDXL
head. 
\end{itemize}

\vspace{-0.75ex}
We note that all four baselines are paired with SDXL as the frozen text-to-image generator, matching the renderer used by \method{}, for fair comparison.

\vspace{-0.5ex}
\subsection{Experiment Settings}
\vspace{-0.5ex}
\paragraph{Models.}
The planner $\pi_\phi$ and reflector $\pi_\psi$ are both initialized
from Qwen2.5-VL-7B. The frozen renderer $G$
is SDXL-1.0 with resolution <1024 $\times$ 1024>. Stage 1 silver trajectories
are produced by a Gemini2.5-Flash teacher with privileged access
to the identity anchor $\mathcal{I}_u^\star$, which is the held out
target image for Personalized Scene Generation and the held out post
for Personalized Creative Generation.

\vspace{-0.5ex}
\paragraph{Training.}
For both stage 1 and 2, we finetune $\pi_\phi$ and $\pi_\psi$
using LoRA adapters~\citep{hu2021loralowrankadaptationlarge}. For stage
2, we sample $K=4$ candidate trajectories from the warm-started policy
with sampling temperature 0.9. For \textit{Personalized Scene
Generation}, we instantiate the reward as a weighted composite of
recommendation rank (reciprocal rank of the target product among
distractors under a CLIP recommender), identity element coverage in the
rendered scene, and a Qwen2.5-VL pointwise judgment. For \textit{Personalized Creative
Generation}, we instantiate the reward as a weighted composite of CLIP
and DINO similarities to the user's held-out post. All experiments run on a single 4090 GPU. 

\begin{table*}[h!]
  \centering
  \caption{Main results on \emph{Personalized Scene Generation} for e-commerce setting on Amazon.
  \\(Bold marks the best per column; underline marks second-best.)}
  \label{tab:task1_main}
  \scriptsize
  \setlength{\tabcolsep}{4pt}
  \begin{tabular}{l cc c ccc}
  \toprule
  & \multicolumn{2}{c}{Retrieval} & Image Quality & \multicolumn{3}{c}{MLLM Judge ($1$--$5$)} \\
  \cmidrule(lr){2-3} \cmidrule(lr){4-4} \cmidrule(lr){5-7}
  Method & H@5\,$\uparrow$ & MRR\,$\uparrow$ & Aes\,$\uparrow$ & Cont.\,$\uparrow$ & Overall\,$\uparrow$ & Sty.\,$\uparrow$ \\
  \midrule
  PMG\,\citep{shen2024pmgpersonalizedmultimodal}            & 0.2143 & \underline{0.1535} & 5.11 & 3.58 & \underline{3.48} & \underline{2.48} \\
  Pigeon\,\citep{xu2025personalized}                        & 0.0985 & 0.0834 & 4.60 & \underline{4.32} & 1.79 & 1.98 \\
  LaVIT\,\citep{jin2024unifiedlanguagevisionpretrainingllm} & \underline{0.2162} & 0.1441 & 4.38 & 3.94 & 2.92 & 2.37 \\
  LLaVA\,\citep{liu2023visualinstructiontuning}             & 0.1873 & 0.1367 & \underline{5.17} & 4.32 & 2.89 & 2.30 \\
  \midrule
  \method{} (ours)                                         & \textbf{0.2297} & \textbf{0.1639} & \textbf{5.33} & \textbf{4.36} & \textbf{3.92} & \textbf{2.88} \\
  \bottomrule
  \end{tabular}
\end{table*}

\vspace{-0.5ex}
\subsection{Quantitative Results}
\vspace{-0.5ex}

\paragraph{Personalized Scene Generation.}
Table~\ref{tab:task1_main} reports results on Personalized Scene
Generation (E-commerce Product Presentation), where \method{} outperforms baselines on all metrics. The largest
margins appear on the personalization-facing metrics such as H@5 and MRR, demonstrating the effective personalization of the proposed framework. Although all baselines share the same image generation backbone, gains in image quality reflect that \method{} produces superior scene reasoning grounded in the user's history. Additionally, the superior performance on the MLLM-based metrics corroborates the gains observed in the personalization and quality metrics, as MLLM judges aggregate style, content, and overall fit into a holistic assessment whose nuance no automatic metric can capture.

\paragraph{Personalized Creative Generation.}
Table~\ref{tab:task2_main} reports results on Personalized Creative Generation (Social Media Content Posting), where \method{} is best on 7 of 10 columns, including all retrieval and MLLM judge metrics. On target-based image metrics, \method{} is strictly best on CIS and DIS, and competitive on the remaining two; notably, the methods that win LPIPS or MS-SSIM score poorly on retrieval, illustrating the value of multi-axis evaluation: a single fidelity metric can be won without delivering personalization. Additionally, the superior performance on the MLLM-based metrics corroborates the gains observed in the personalization and quality metrics.

\begin{table*}[t]
  \centering
  \caption{Main results on \emph{Personalized Creative Generation} for social media setting on Instagram. \\ (Bold marks the best per
   column; underline marks second-best.)}
  \label{tab:task2_main}
  \scriptsize
  \setlength{\tabcolsep}{1.9pt}
  \begin{tabular}{@{}l ccccc cc ccc@{}}
  \toprule
  & \multicolumn{5}{c}{Target-based Image Metrics} & \multicolumn{2}{c}{Contrastive R@1} &
  \multicolumn{3}{c}{MLLM Judge ($1$--$5$)} \\
  \cmidrule(lr){2-6} \cmidrule(lr){7-8} \cmidrule(lr){9-11}
  Method & CIS\,$\uparrow$ & DIS\,$\uparrow$ & LPIPS\,$\downarrow$ & MS-SSIM\,$\uparrow$ &
  Aes.\,$\uparrow$ & Inter-cat.\,$\uparrow$ & Intra-cat.\,$\uparrow$ & Sty.\,$\uparrow$ &
  Cont.\,$\uparrow$ & Overall\,$\uparrow$ \\
  \midrule
  \shortstack[l]{PMG\\\citep{shen2024pmgpersonalizedmultimodal}}                          & 0.601 & 0.206 &
  \textbf{0.729} & 0.047 & \textbf{5.84} & 0.597 & 0.466 & 3.167 & 3.075 & 3.157 \\
  \shortstack[l]{Pigeon\\\citep{xu2025personalized}}                                       & 0.561 &
  \underline{0.256} & 0.741 & 0.053 & 4.28 & \underline{0.748} & \underline{0.659} & 3.341 &
  \underline{3.258} & \underline{3.334} \\
  \shortstack[l]{LaVIT\\\citep{jin2024unifiedlanguagevisionpretrainingllm}}               & \underline{0.616} & 0.238
   & 0.744 & \textbf{0.068} & 5.37 & 0.640 & 0.395 & \underline{3.344} & 3.143 & 3.331 \\
  \shortstack[l]{LLaVA\\\citep{liu2023visualinstructiontuning}}                           & 0.575 & 0.171 &
  \underline{0.735} & 0.053 & \underline{5.48} & 0.432 & 0.285 & 3.256 & 3.179 & 3.227 \\
  \midrule
  \method{} (ours)                                                       & \textbf{0.645} &
  \textbf{0.289} & \underline{0.735} & \underline{0.055} & 5.45 & \textbf{0.876} & \textbf{0.701} &
  \textbf{3.554} & \textbf{3.277} & \textbf{3.541} \\
  \bottomrule
  \end{tabular}
  \end{table*}

\vspace{-0.5ex}
\subsection{Human Evaluation}
\label{sec:human-evaluation}
\vspace{-0.5ex}
We complement automatic evaluation with the initial SDXL human study, in which one annotator assessed 50 distinct test users: 25 scene-generation pairs against PMG and 25 creative-generation pairs against Pigeon. Method names were hidden, case order was randomized, and left/right placement was balanced within each task. For fit to the displayed user history, the annotator selected \method{} in 19/25 scene pairs (76\%) and PMG in 4/25 (16\%), with insufficient evidence in two cases. For creative generation, \method{} was selected in 16/25 pairs (64\%) and Pigeon in 9/25 (36\%). These are preliminary, single-annotator judgments of history fit, not the target users' own preferences. Appendix~\ref{app:human-study} provides the sampling protocol, all response categories, and conditioning limitations.

\vspace{-0.5ex}
\subsection{Qualitative Results}
\begin{figure}[h]
\centering
\begin{subfigure}[t]{0.35\textwidth}
\centering
\includegraphics[width=\linewidth]{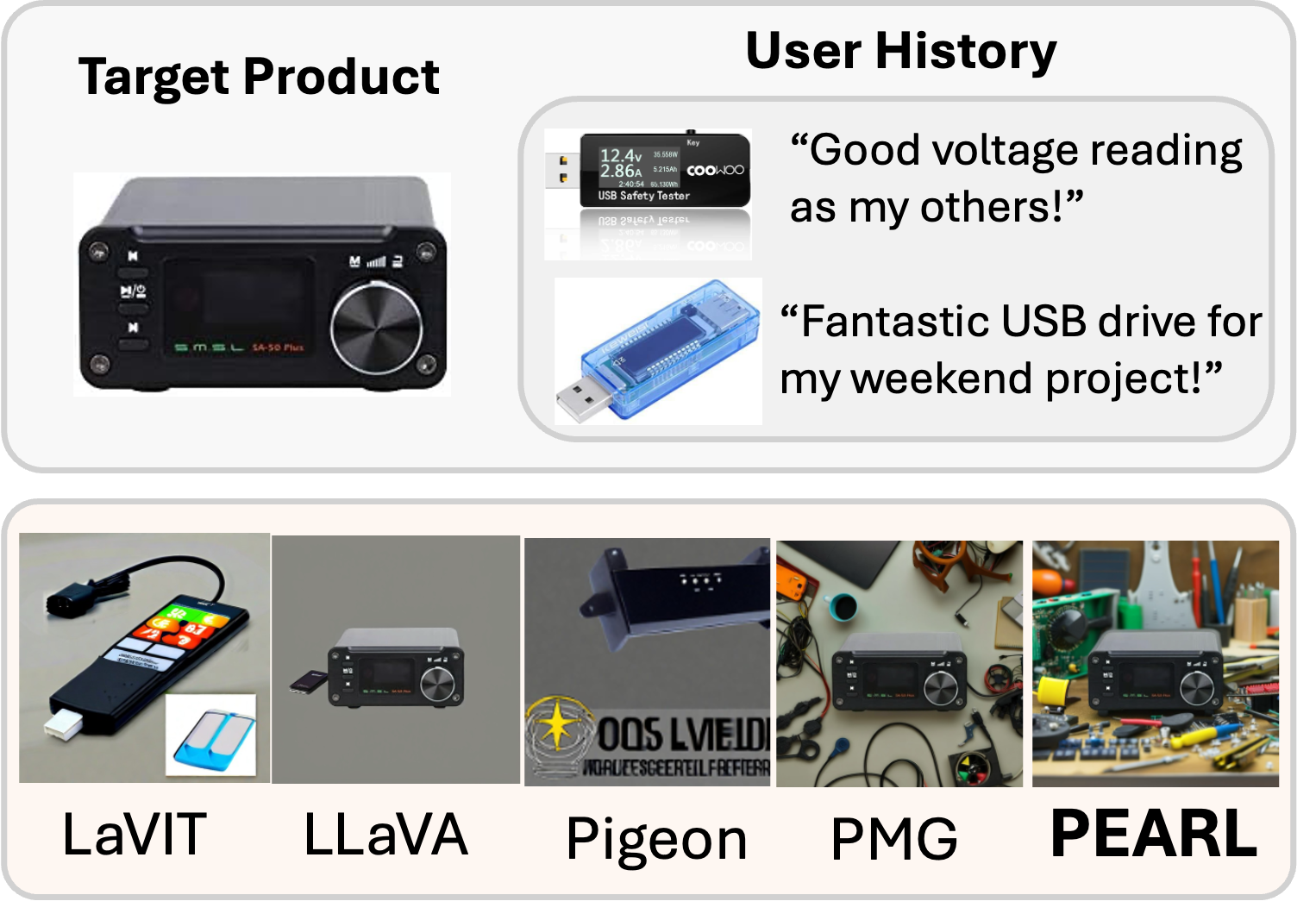}
\caption{Personalized Scene Generation.}
\label{fig:psg}
\end{subfigure}
\hspace{0.05\textwidth}
\begin{subfigure}[t]{0.35\textwidth}
\centering
\includegraphics[width=\linewidth]{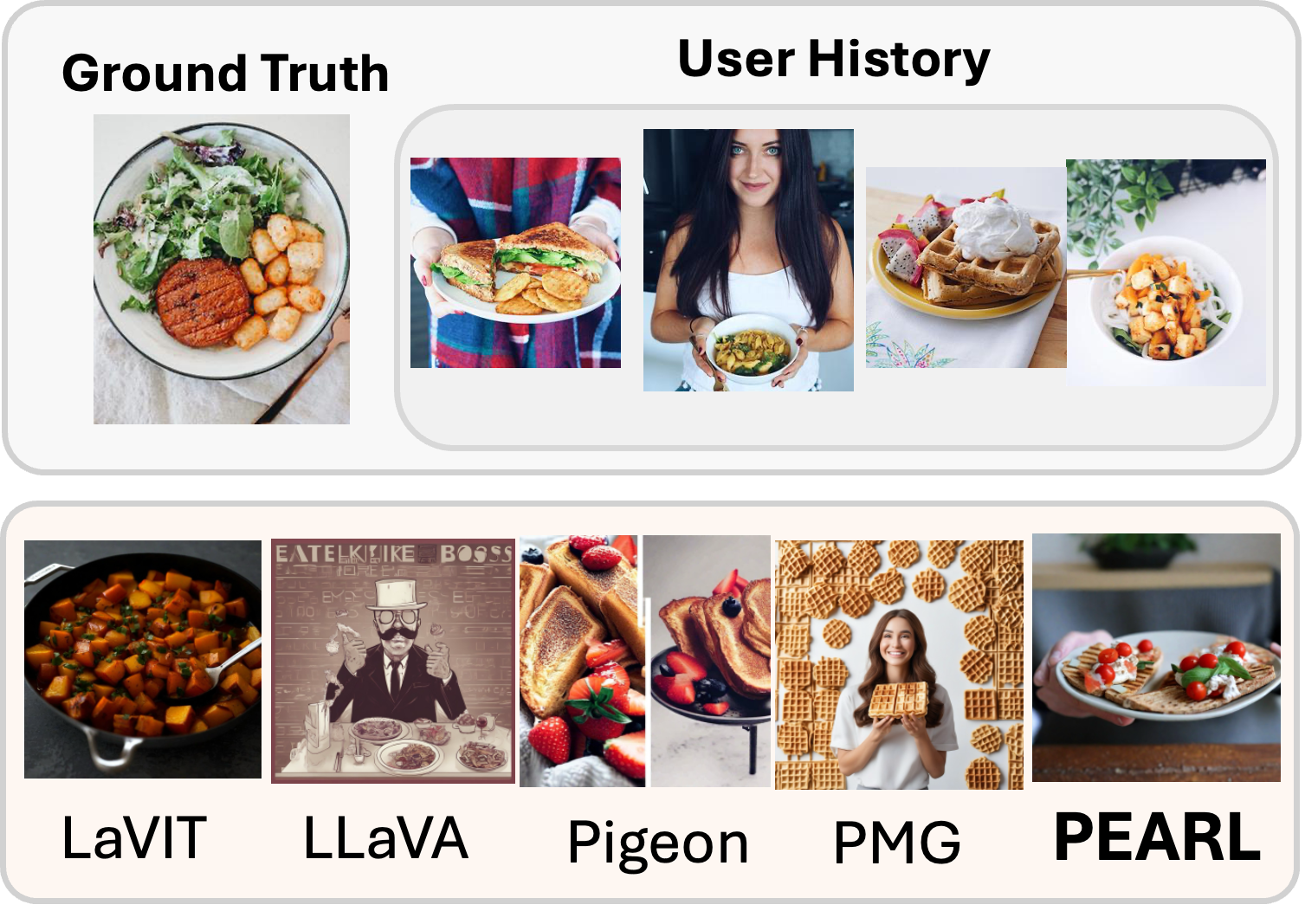}
\caption{Personalized Creative Generation.}
\label{fig:pcg}
\end{subfigure}
\caption{Qualitative Results.}
\label{fig:qualitative}
\vspace{2ex}
\end{figure}

\vspace{-0.5ex}
We include the qualitative results in Figure~\ref{fig:qualitative}. In Figure~\ref{fig:psg}, Compared to the baselines, \method successfully captures the user identity from the user history, which entails a weekend DIY enthusiast with a working table. LaVIT and Pigeon, whose personalization requires conditioned generation on the embeddings, fails to preserve the product detail and generate meaningful images from the user history. Figure~\ref{fig:pcg} shares the same story where \method captures the details of the user aesthetic, where the food is often presented in clean, light-colored kitchenware with a hand offering gesture. Together, these examples illustrate how \method translates user history into concrete choices about scene context, composition, and visual presentation. We present more qualitative results in Appendix~\ref{app:qualitative}. 

\subsection{Ablation Studies}
\vspace{-0.5ex}
\ifdefined\arxivpreprint
\begin{figure}[t]
\else
\begin{wrapfigure}{r}{0.35\textwidth}
\fi
\centering
\ifdefined\arxivpreprint
\includegraphics[width=0.36\linewidth]{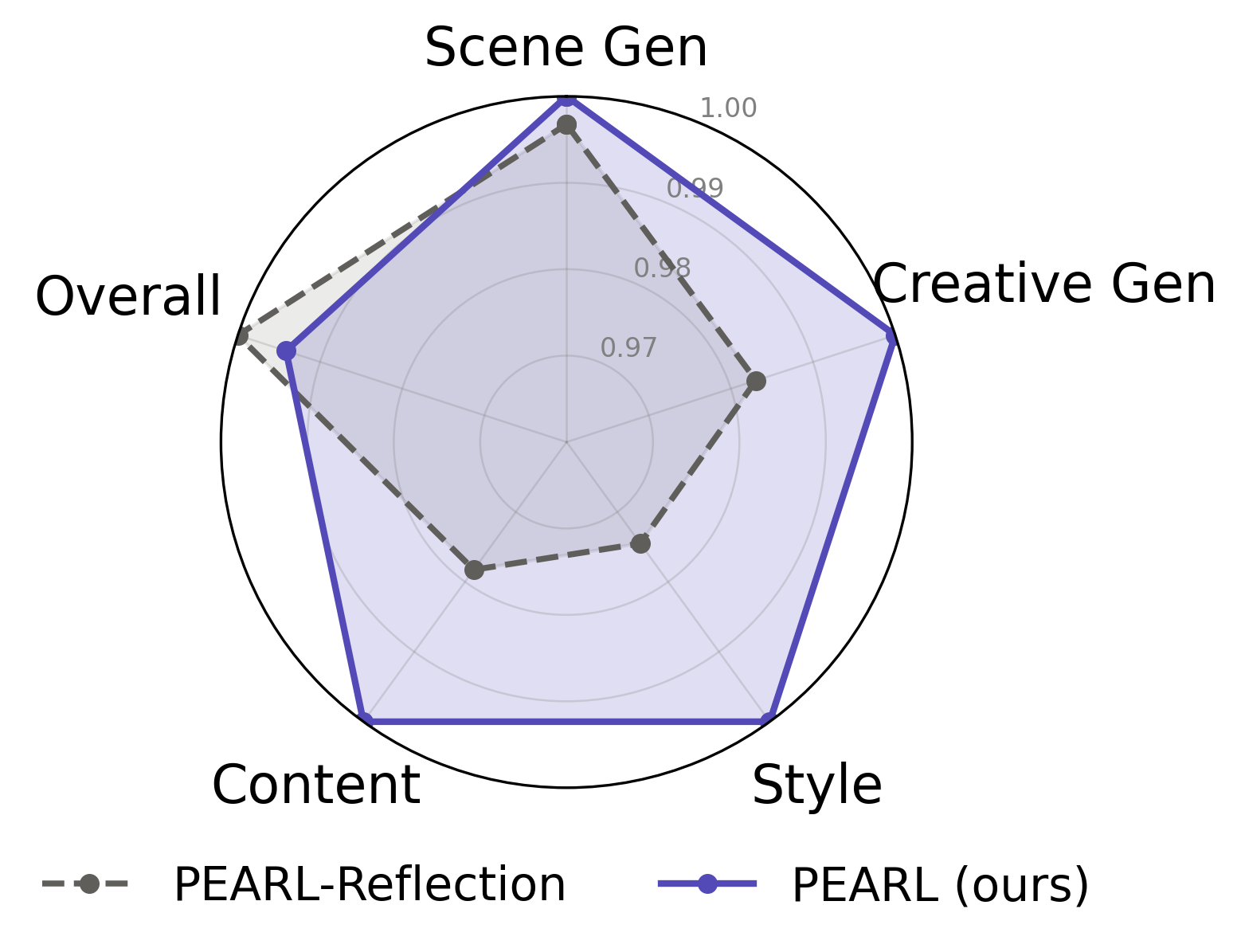}
\else
\vspace{-12pt}
\includegraphics[width=\linewidth]{imgs/ablation_radar.png}
\fi
\caption{\textbf{Ablation: \method{} vs.\ \method{}-Reflection.} The visualization normalizes and aggregates the metrics across both tasks.}
\label{fig:ablation-radar}
\ifdefined\arxivpreprint
\end{figure}
\else
\vspace{-8pt}
\end{wrapfigure}
\fi
We further investigate the effectiveness of the reflector module by comparing the full \method{} against \method{}-Reflection, an ablation in which we replace the alternating-policy DPO stage with supervised fine-tuning alone. We present the results in Figure~\ref{fig:ablation-radar}. Across both \dataset{} tasks and the MLLM judge, \method{} outperforms \method{}-Reflection on four of the five aggregated dimensions, demonstrating that closing the loop between reasoning and rendering through preference optimization yields measurable gains over supervised distillation alone. We note, however, that the largest improvements appear on the MLLM judge metrics, particularly style and content, suggesting that the reflector primarily contributes by aligning fine-grained details that surface-level retrieval and image-quality metrics do not fully capture. The marginal drop on overall score (under $1\%$) suggests that the gains from the reflector concentrate on fine-grained personalization details rather than on drastic adjustments. This is consistent with the findings of IRG~\citep{huang2025interleavingreasoningbettertexttoimage}, which similarly observed that interleaved reasoning yields its largest gains on fine-grained fidelity rather than on coarse semantic alignment. The detailed per-metric results are provided in Appendix~\ref{app:ablation}.

\section{Related Work}\label{sec:related-work}
\paragraph{Conditional Personalized Image Generation.}
Conditional personalized image generation~\citep{wei2025personalizedimagegenerationdeep} synthesizes images incorporating a user-specified visual concept under a novel text prompt, given a small set of reference images depicting it. Existing methods fall into two families. \emph{Test-time optimization} methods such as DreamBooth~\citep{ruiz2023dreamboothfinetuningtexttoimage} and Textual Inversion~\citep{yang2023controllabletextualinversionpersonalized} fine-tune the generator or learn a dedicated text embedding to bind the reference concept, achieving strong subject fidelity but requiring per-user training. \emph{Adapter-based} methods such as IP-Adapter~\citep{ye2023ipadaptertextcompatibleimage} amortize this cost by encoding reference images directly into the cross-attention conditioning of a frozen generator. In both families, ``personalization'' means faithfully reproducing a pre-specified visual concept. Our work instead infers a user's identity from realistic histories and synthesizes scenes that reflect it.

\paragraph{History-Based Personalized Image Generation.}
Recent work conditions generation on a user's accumulated interaction history rather than curated reference images. Pigeon~\citep{xu2025personalized} uses a large multimodal model with dedicated masking and aggregation modules to encode noisy user histories alongside multimodal instructions, and is evaluated on personalized sticker and movie poster generation. PMG~\citep{shen2024pmgpersonalizedmultimodal} instead translates history images into textual descriptions and uses a large language model to encode user preferences from this textual proxy, trading visual fidelity for the flexibility of language-based reasoning. Personalized Image Editing~\citep{dunlop2025personalized} takes an editing approach, using Collaborative DPO over a learned preference graph to share aesthetic signals across users with similar tastes. Although these methods establish user history as a useful conditioning signal, they fuse it directly into the generator without explicit reasoning over its implications for the target image. \method{} instead reasons explicitly over multimodal user histories.

\section{Conclusion}
\label{sec:conclusion}
We introduced \dataset, the first benchmark for personalized image generation grounded in realistic, multimodal user histories from public review and posting platforms, and \method, a reasoning-interleaved framework that decomposes the problem into identity reasoning over the user's history and conditioned synthesis through a frozen image generator. \method{} optimizes its \emph{render-then-reflect} loop through a two-stage procedure that combines silver-trajectory supervision with alternating preference optimization. Across both tasks on \dataset, \method{} substantially outperforms strong personalization baselines on image quality, retrieval-based personalization, and MLLM-judge metrics, and ablations confirm that both reasoning and reflection contribute measurable gains. 

\section{AI Use Disclosure}
In this work, we used generative AI tools for brainstorming research idea, implementing part of the codebase, and do preliminary analysis on the experiment result. We have not used generative AI tools for autoresearch, propose hypothesis, or provide feedback on methodology, and the rest of the required disclosure tasks are not applicable to this work. Additionally, we used generative AI tools for draft part of the research paper, create or modify the teaser figure and images, and summarize existing landscape in the research area that is within our interests. We have reviewed all AI-assisted work. We checked LLM-generated research ideas for potential plagiarism through a manual literature survey, and audited the claims and code written by LLMs. We take responsibility for the final content of this work, including text, claims or artifacts produced with the aid of generative AI.

\bibliography{Styles/paper}
\bibliographystyle{iclr2026_conference}

\clearpage
\appendix
\newpage

\section{Limitations}
\label{sec:limitations}

Our work has two main limitations. First, Personalized Scene Generation has no naturally occurring ground-truth scene image per (user, product) pair, so we evaluate user alignment indirectly through retrieval-based and judge-based metrics rather than direct reference-based comparison. Second, our \method{} training procedure performs only a single round of alternating-policy DPO updates between the planner and the reflector; while we expect additional rounds to yield further gains, a systematic study of multi-round optimization remains future work.

\section{\dataset}
\label{app:dataset}

In this section, we provide additional details on the two task instantiations in our personalized image generation benchmark: Personalized Scene Generation for e-commerce product presentation and Personalized Creative Generation for social media content posting. While the main paper defines the common problem formulation, here we describe the task motivations, source corpora, filtering procedures, target construction, and user-level splits used to construct each dataset.

The two tasks are designed to evaluate complementary forms of user-history-based personalization. Personalized Scene Generation focuses on adapting the visual context around a product to an individual user's inferred interests and lifestyle, motivated by industrial settings in which platforms or sellers may personalize product imagery to increase engagement, clicks, and purchase intent. Personalized Creative Generation focuses on extending a user's visual identity to new content, motivated by creator-support workflows in which tools help influencers decide what to post and how to present it while preserving their long-standing style.

In both tasks, the generation condition specifies the content to be rendered, while the user history provides the personalization signal. For scene generation, the condition is the product image to be personalized; for creative generation, the condition is a topic or content specification. This separation allows us to evaluate whether models can go beyond generic conditional generation by adapting the output to user-specific behavioral or aesthetic patterns inferred from history.

Table~\ref{tab:dataset_stats} summarizes the resulting dataset statistics across tasks and splits. We next describe the construction of each task in detail, beginning with the e-commerce scene-generation setting and then turning to the social media creative-generation setting.

\begin{table}[h]
\centering
\caption{Dataset statistics.}
\label{tab:dataset_stats}
\begin{tabular}{l|cc|ccc}
\toprule
& \multicolumn{2}{c|}{\shortstack{Personalized Scene Generation\\(E-commerce: Amazon)}} & \multicolumn{3}{c}{\shortstack{Personalized Creative Generation\\(Social Media: Instagram)}} \\
\cmidrule(lr){2-3} \cmidrule(lr){4-6}
Split & \#Users & \#Reviews & \#Users & \#Posts & \#Pairs \\
\midrule
Train & 3,972 & 39,720 & 3,464 & ~40K & 957 \\
Val   & 510   & 5,100  & 742   & ~8K  & --- \\
Test  & 518   & 5,180  & 207   & ~2K  & 314 \\
\midrule
Total & 5,000 & 50,000 & 4,413 & 50,656 & 1,271 \\
\bottomrule
\end{tabular}
\end{table}

\subsection{Personalized Scene Generation: E-commerce Product Presentation}
\label{app:scene-data}

\paragraph{Motivation. }
This task is motivated by the practical goal of improving user engagement in e-commerce through personalized product imagery. Prior work has shown that users tend to respond more favorably to personalized images, for example through higher click-through rates, although the little work that does exist in the literature has primarily studied this at a coarse category or segment level. Here, we formulate a more fine-grained benchmark from the perspective of a platform such as Amazon, or a seller operating on such a platform, where the objective is to generate product images tailored to an individual user's inferred interests and lifestyle. The product itself should remain faithful, but the surrounding visual context should be adapted so that it is more likely to attract attention, encourage clicks, and ultimately increase purchase intent. In this sense, the task serves as an offline benchmark for personalized product presentation and generative recommendation at the user level.

\vspace{-0.75ex}
\paragraph{Source corpus.}
We build the Personalized Scene Generation dataset on top of the Amazon Reviews 2023 corpus~\citep{hou2024bridging}, which spans 33 product categories with timestamped review text, ratings, product metadata, and user-uploaded review images. We restrict our pool to six categories that exhibit strong lifestyle signal (\textit{Sports \& Outdoors}, \textit{Home \& Kitchen}, \textit{Tools \& Home Improvement}, \textit{Automotive}, \textit{Pet Supplies}, and \textit{Electronics}); these categories together cover the principal axes of consumer behavior we want the dataset to probe.

\vspace{-0.75ex}
\paragraph{User filtering.}
A history is informative only if it contains enough textual signal to support identity inference. We adopt the LongLaMP~\citep{kumar2024longlampbenchmarkpersonalizedlongform} density filter: a user is retained only if they have at least four reviews of length at least 120 words. To further ensure each history is non-trivially diverse, we additionally require that a user's reviews span at least three of the six categories, which filters out users whose interests are too narrow to test cross-category lifestyle inference. We strip HTML tags, deduplicate reviews by hash, and discard reviews flagged as non-English by fastText language identification.

\vspace{-0.75ex}
\paragraph{Target construction.}
For each user, we hold out one review (uniformly at random from those satisfying the density filter) as the generation target. The remaining reviews form the user's history $\mathcal{H}_u$, and the held-out review's product image becomes the the generation condition $c=o^\star$. Drawing target products from each user's own history (rather than synthetically pairing arbitrary products with arbitrary users) preserves the natural user-product affinity present in real e-commerce: the model is asked to render a product the user has actually engaged with, not an out-of-distribution match.

\vspace{-0.75ex}
\paragraph{Splits.}
We partition users into train / validation / test splits of approximately 80\,:\,10\,:\,10. Splits are user-disjoint, meaning a user's reviews appear in only one split, so generalization to unseen users is tested directly. 

\subsection{Personalized Creative Generation: Social Media Content Posting}
  \label{app:creative-data}

\paragraph{Motivation. }
This task is motivated by the need for tools that help creators plan content while maintaining a consistent visual identity. Influencers often rely on consultants to advise what to post, when to post it, and how to present it so that new content aligns with their long-standing style. However, a creator's style is not always reducible to explicit promptable attributes such as ``minimalist,'' ``bright,'' or ``vintage.'' Many stylistic cues are implicit, habitual, or even subconscious, such as repeatedly photographing food from a top-down angle, placing products near natural light, centering pets in domestic spaces, using muted backgrounds with a single saturated object, favoring close-up hand-held compositions, or pairing travel scenes with wide negative space. We formulate Personalized Creative Generation as a benchmark for inferring these latent aesthetic regularities from the user's visual history and applying them to a new topic. The goal is not to inject a few user-provided style keywords into a generic image generator, but to generate content that could plausibly fit into the user's existing feed while depicting the requested topic.

\vspace{-0.75ex}
  \paragraph{Source corpus.}
  We build the Personalized Creative Generation dataset on top of the
  Instagram Influencer Dataset~\citep{kim2020multimodal}, a
  publicly released collection of 33{,}935 influencers covering nine topical
  categories (\emph{beauty, family, fashion, fitness, food, interior, pet,
  travel, other}). For each influencer the dataset provides up to 300
  chronologically ordered posts, each consisting of one image, a free-text
  caption, hashtags, user-tags, a timestamp, and engagement metadata
  (likes, comments). We restrict our pool to five lifestyle-rich categories
  (\emph{food, interior, pet, travel, other}) for the held-out evaluation, as
  these categories carry the strongest visual style signal and the smallest
  overlap with celebrity / promotional content present in
  \emph{fashion}, \emph{beauty}, and \emph{family}. The training and
  validation pools retain all nine categories so the reasoning model is
  exposed to a broader stylistic distribution during learning.

\vspace{-0.75ex}
  \paragraph{User filtering.}
  A history is informative only if it spans enough chronologically distinct
  posts to expose a stable visual identity. We retain a user only if they
  have at least four posts on record, drop posts marked as videos, and drop
  posts whose stored image cannot be loaded (corrupted or removed). We
  discard users whose posts are dominated by sponsored content
  (\texttt{is\_ad}{=}\texttt{True} for more than half their posts) to avoid
  fitting commercial templates rather than personal style. Finally, posts are
  deduplicated by image hash to remove the cross-posted duplicates that are
  common among influencer accounts.

\vspace{-0.75ex}
  \paragraph{Target construction.}
  For each user, we sort posts chronologically by timestamp. The two most
  recent posts are held out as generation targets, and the preceding five
  posts are used as the visual history $\mathcal{H}_u$ presented to the
  reasoning model. This temporal split mirrors real social media,
  where a creator's next post must be predicted from past activity rather
  than from a randomly held-out window. 

\vspace{-0.75ex}
  \paragraph{Splits.}
  We partition users into train / validation / test splits of approximately
  $70\,{:}\,15\,{:}\,15$ at the user level, so a user's posts appear in only
  one split and generalization to unseen users is tested directly. The
  resulting corpus contains $3{,}464$ training users
  ($\sim$$40\mathrm{K}$ posts), $742$ validation users
  ($\sim$$8\mathrm{K}$ posts), and $207$ test users
  ($\sim$$2\mathrm{K}$ posts), with the test split restricted to the five
  lifestyle categories above. After applying the chronological history
  window we obtain $957$ training (user, target) pairs and $314$ test
  (user, target) pairs used in all reported experiments.

\section{\method{} Algorithms}
\label{app:algorithms}

We provide pseudocode for the full \method{} pipeline at inference time and for the two-stage training procedure in Algorithms~\ref{alg:pearl} and~\ref{alg:pearl-training}, respectively. The training algorithm formalizes the two-stage procedure described in Section~\ref{sec:method-train}: Stage~1 distills silver trajectories from a teacher model with privileged access to an identity signal $\mathcal{I}_u^\star$, fine-tuning the planner and reflector independently via cross-entropy on the resulting trajectories; Stage~2 alternates between updating the two policies via Direct Preference Optimization, where each policy's candidate outputs are scored by running them through the full downstream pipeline (including the other, frozen policy and the renderer) and ranked by a task-aligned retrieval reward $R$. 
\begin{algorithm}[t]
\caption{\method{} Framework}
\label{alg:pearl}
\small
\begin{algorithmic}[1]
\Require User history $\mathcal{H}_u = \{h^{(u)}_i\}_{i=1}^{N_u}$, generation condition $c$, frozen text-to-image generator $G$, trained planner $\pi_\phi$, trained reflector $\pi_\psi$
\Ensure Personalized image $\hat{I}$
\Statex
\Statex \textit{\textbf{Stage 1: Planning}} \Comment{Section~\ref{sec:method-planner}}
\State Construct planner prompt from $\mathcal{H}_u$ and $c$
\State Sample reasoning trace: $r_1 \sim \pi_\phi(\cdot \mid \mathcal{H}_u, c)$ \Comment{Summarize user lifestyle, aesthetics, behavioral patterns}
\State Sample scene description: $s_1 \sim \pi_\phi(\cdot \mid \mathcal{H}_u, c, r_1)$ \Comment{Self-contained prompt for $G$}
\Statex
\Statex \textit{\textbf{Initial Rendering}}
\State $I_1 \leftarrow G(s_1, c)$ \Comment{Render planner output via frozen generator}
\Statex
\Statex \textit{\textbf{Stage 2: Reflection}} \Comment{Section~\ref{sec:method-reflector}}
\State Construct reflector prompt from $\mathcal{H}_u$, $I_1$, and $c$
\State Sample structured analysis: $r_2 \sim \pi_\psi(\cdot \mid \mathcal{H}_u, I_1, c)$
\Statex \hspace{\algorithmicindent} \Comment{Compare $I_1$ against $\mathcal{H}_u$: scene consistency, lifestyle cues, aesthetic alignment}
\State Sample revised scene description: $s_2 \sim \pi_\psi(\cdot \mid \mathcal{H}_u, I_1, c, r_2)$ \Comment{Corrected prompt for $G$}
\Statex
\Statex \textit{\textbf{Final Rendering}} \Comment{Eq.~\ref{eq:reflector-render}}
\State $\hat{I} \leftarrow G(s_2, c)$ \Comment{Re-render with refined plan}
\Statex
\State \Return $\hat{I}$, $(r_1, s_1)$, $(r_2, s_2)$
\end{algorithmic}
\end{algorithm}

\begin{algorithm}[t]
\caption{\method{} Training}
\label{alg:pearl-training}
\begin{algorithmic}[1]
\Require Training set $\mathcal{D} = \{(\mathcal{H}_u, c, \mathcal{I}_u^\star)\}$, frozen generator $G$, teacher model $\mathcal{T}$, retrieval reward $R$, number of candidates $K$, DPO temperature $\beta$
\Ensure Trained planner $\pi_\phi$, trained reflector $\pi_\psi$
\Statex
\Statex \textit{\textbf{Stage 1: Silver Trajectory Distillation}} \Comment{Section~\ref{sec:method-train}}
\For{each $(\mathcal{H}_u, c, \mathcal{I}_u^\star) \in \mathcal{D}$}
    \State $(r_1^\star, s_1^\star) \sim \mathcal{T}(\cdot \mid \mathcal{H}_u, c, \mathcal{I}_u^\star)$ \Comment{Silver planner trajectory}
    \State $\tilde{I}_1 \leftarrow G(s_1^\star, c)$ \Comment{Render pseudo-initial image}
    \State $(r_2^\star, s_2^\star) \sim \mathcal{T}(\cdot \mid \mathcal{H}_u, \tilde{I}_1, c, \mathcal{I}_u^\star)$ \Comment{Silver reflector trajectory}
\EndFor
\State $\pi_\phi \leftarrow \arg\min_\phi \; \mathcal{L}_{\text{SFT}}(\pi_\phi)$ \Comment{Fine-tune planner}
\State $\pi_\psi \leftarrow \arg\min_\psi \; \mathcal{L}_{\text{SFT}}(\pi_\psi)$ \Comment{Fine-tune reflector}
\State $\pi_\phi^{\text{ref}} \leftarrow \pi_\phi, \quad \pi_\psi^{\text{ref}} \leftarrow \pi_\psi$ \Comment{Store reference policies for DPO}
\Statex
\Statex \textit{\textbf{Stage 2: Render-in-the-Loop Preference Optimization}}
\Statex
\Statex \textit{Planner Update} \Comment{Reflector $\pi_\psi$ and $G$ frozen}
\For{each $(\mathcal{H}_u, c) \in \mathcal{D}$}
    \For{$k = 1, \dots, K$}
        \State $(r_1^{(k)}, s_1^{(k)}) \sim \pi_\phi(\cdot \mid \mathcal{H}_u, c)$ \Comment{Sample planner candidate}
        \State $I_1^{(k)} \leftarrow G(s_1^{(k)}, c)$ \Comment{Render initial image}
        \State $(r_2^{(k)}, s_2^{(k)}) \sim \pi_\psi(\cdot \mid \mathcal{H}_u, I_1^{(k)}, c)$ \Comment{Run frozen reflector}
        \State $\hat{I}^{(k)} \leftarrow G(s_2^{(k)}, c)$ \Comment{Render final image}
        \State $\alpha^{(k)} \leftarrow R(\hat{I}^{(k)}, \mathcal{H}_u)$ \Comment{Score via retrieval reward}
    \EndFor
    \State $(r_1^w, s_1^w) \succ (r_1^l, s_1^l)$ from $\arg\max_k \alpha^{(k)}$ and $\arg\min_k \alpha^{(k)}$ \Comment{Form preference pair}
\EndFor
\State $\pi_\phi \leftarrow \pi_\phi - \eta \nabla_\phi \, \mathcal{L}_{\text{DPO}}(\pi_\phi)$ \Comment{Update planner (Eq.~\ref{eq:dpo})}
\Statex
\Statex \textit{Reflector Update} \Comment{Planner $\pi_\phi$ and $G$ frozen}
\For{each $(\mathcal{H}_u, c) \in \mathcal{D}$}
    \State $(r_1, s_1) \sim \pi_\phi(\cdot \mid \mathcal{H}_u, c)$; \quad $I_1 \leftarrow G(s_1, c)$ \Comment{Generate input from frozen planner}
    \For{$j = 1, \dots, K$}
        \State $(r_2^{(j)}, s_2^{(j)}) \sim \pi_\psi(\cdot \mid \mathcal{H}_u, I_1, c)$ \Comment{Sample reflector candidate}
        \State $\hat{I}^{(j)} \leftarrow G(s_2^{(j)}, c)$ \Comment{Render refined image}
        \State $\gamma^{(j)} \leftarrow R(\hat{I}^{(j)}, \mathcal{H}_u)$ \Comment{Score via retrieval reward}
    \EndFor
    \State $(r_2^w, s_2^w) \succ (r_2^l, s_2^l)$ from $\arg\max_j \gamma^{(j)}$ and $\arg\min_j \gamma^{(j)}$ \Comment{Form preference pair}
\EndFor
\State $\pi_\psi \leftarrow \pi_\psi - \eta \nabla_\psi \, \mathcal{L}_{\text{DPO}}(\pi_\psi)$ \Comment{Update reflector (Eq.~\ref{eq:dpo})}
\Statex
\State \Return $\pi_\phi$, $\pi_\psi$
\end{algorithmic}
\end{algorithm}

\section{Evaluation Protocol Details}
\label{app:evaluation}

This appendix expands on the metrics summarized in Section~\ref{sec:eval}, providing precise definitions, construction details, and prompts.

\subsection{Contrastive Retrieval Probes: Shared Setup}
\label{app:probe-shared}

Both \dataset{} tasks include a contrastive retrieval probe that scores how well a generated image places its user in their natural retrieval pool. The probes share the following protocol; we describe the task-specific user encoders and pool constructions in Sections~\ref{app:eval-reference} and~\ref{app:contrastive}.

\paragraph{Probe formulation.} A learned compatibility scorer $s(u, I)$ assigns a score to each $(\text{user}, \text{image})$ pair. The scorer is trained on real $(\text{history}, \text{target})$ pairs from the training split with InfoNCE over in-batch negatives, then frozen for evaluation. At evaluation we substitute the real target with the variant's \emph{generated} image $\hat{I}_u$ and measure the resulting rank against a deterministic distractor pool (seeded by $\mathrm{hash}(u)$, identical across methods). Generated images never enter the training distribution, so there is no circularity in using the probe to evaluate generations.

\paragraph{Visual encoding.} For both probes, images are encoded with frozen CLIP\,ViT-L/14 (OpenAI weights) into a $768$-d feature, then projected through a learned MLP head ($768 \to 512 \to 256$ with GELU, LayerNorm, and dropout $0.1$) and $\ell_2$-normalized. Compatibility is a temperature-scaled cosine, $s(u, I) = \tau^{-1}\langle \mathbf{u}, \mathbf{e}_I\rangle$, with $\tau$ a learned scalar.

\paragraph{Metrics.} Given the rank $r(\hat{I}_u)$ of the true target (or correct user) among the candidates,
\[
\mathrm{Hit@}K = \frac{1}{|\mathcal{U}|}\sum_u \mathbf{1}[r(\hat{I}_u) \leq K], \qquad
\mathrm{MRR} = \frac{1}{|\mathcal{U}|}\sum_u \frac{1}{r(\hat{I}_u)}.
\]

\subsection{Recommendation Probe for Personalized Scene Generation}
\label{app:eval-reference}

The recommendation intuition is that a personalized image should make a target product more discoverable to the user it was generated for than a generic catalog photo would. The user encoder is bimodal because Amazon histories carry both behavioral (purchase) and intent (review-text) signals.

\paragraph{Bimodal user encoder.} Each user's history consists of $H{=}9$ items, each pairing a catalog photo and the corresponding review text. The visual stream uses the shared CLIP encoder of Section~\ref{app:probe-shared}; the textual stream encodes each review (truncated to $2{,}000$ characters) with sentence-transformers \texttt{all-mpnet-base-v2} into a $768$-d feature $\mathbf{t}_i$. Per-item bimodal features are concatenated and projected to a per-item embedding $\mathbf{e}_i = \mathrm{MLP}([\mathbf{v}_i;\mathbf{t}_i]) \in \mathbb{R}^{256}$, then masked-attention-pooled into the user embedding $\mathbf{u} \in \mathbb{R}^{256}$. Including review text disambiguates lifestyle signals that purchase photos alone cannot convey (e.g., a user who buys both hiking boots and dress shoes; the reviews disambiguate which is the recurring theme).

\paragraph{Training.} Trained on the $4{,}000$ training-split users with batch size $128$, AdamW (lr $1{\times}10^{-3}$, weight decay $1{\times}10^{-4}$), feature dropout $0.1$ (random history-item drop), and $30$ epochs with early stopping on validation Hit@1.

\paragraph{Distractor pool and ranking.} For each test user with target product $p^\star$, the candidate set contains $p^\star$ plus $99$ distractor products sampled from the full $43{,}834$-product catalog (excluding $p^\star$). At evaluation, $p^\star$'s CLIP feature is replaced with the CLIP feature of $\hat{I}_u$ and all $100$ candidates are scored against the user embedding. We report Hit@5 and MRR over the $518$ test users.

\subsection{Style-Discrimination Probe for Personalized Creative Generation}
\label{app:contrastive}

The style-discrimination intuition is that a strong personalized image should look distinctively like \emph{this} user's posts rather than a generic post in the same topic category. Generic-looking generations are easy to confuse with other users' posts and provide little value to the creator workflow.

\paragraph{Visual-only user encoder.} Unlike Amazon reviews, Instagram captions are short, often non-English, and frequently just hashtags, so the textual stream contributes little reliable identity signal. We therefore encode histories visually only. Each user's $H{=}20$ most recent posts are encoded through the shared CLIP encoder of Section~\ref{app:probe-shared} and attention-pooled with a learned scalar head into $\mathbf{u} \in \mathbb{R}^{256}$.

\paragraph{Training.} Trained on the $3{,}464$ training-split users with batch size $32$, AdamW (lr $1{\times}10^{-4}$, weight decay $1{\times}10^{-2}$), $2$-epoch warm-up, history dropout $0.1$, and $50$ epochs with early stopping on validation Hit@1.

\paragraph{Negative pool difficulty.} Social media style discrimination has a distinctive structure: similar-category users (two food bloggers) can be visually very close, while cross-category users (food vs.\ travel) are trivially separable. To probe both regimes, we construct three negative pools per test user:
\begin{itemize}
    \item \textbf{Inter-category (easy):} $10$ distractor users from \emph{different} topic categories (chance $1/11 \approx 0.091$).
    \item \textbf{Intra-category (medium):} $5$ distractor users from the \emph{same} category but with different visual style, identified as the bottom $50\%$ of users by cosine similarity of their CLIP history centroid to $u$'s (chance $1/6 \approx 0.167$).
    \item \textbf{Intra-style (hard):} $3$ distractor users from the same category whose history centroids are closest to $u$'s (top $5\%$ by cosine similarity, chance $1/4 = 0.250$).
\end{itemize}

\paragraph{Ranking.} For each test user $u$ with target post $p^\star$, the candidate set contains $u$ plus the appropriate distractor pool. We replace $p^\star$ with $\hat{I}_u$, encode it through the frozen image branch, and rank all candidate users by $s(u', \hat{I}_u)$. We report R@1 at the inter-category and intra-category difficulty levels over the $314$ test (user, target) pairs.

\subsection{Output Quality}
\label{app:eval-quality}

We use the LAION aesthetic predictor (the LAION-Aesthetics V2 model~\citep{schuhmann2022laion5bopenlargescaledataset}) to score each generation on a continuous scale roughly $1$--$10$, with higher scores corresponding to images judged more aesthetically pleasing on the LAION training distribution. This is a reference-free metric that verifies improvements on fidelity or user alignment do not come at the cost of obvious image quality regressions. We report the mean over test instances.

\subsection{MLLM-as-Judge Protocol}
\label{app:eval-mllm}

We elicit holistic user-alignment judgments via an ensemble of two multimodal LLMs: Gemini~Flash~2.5 and Qwen2.5-VL-32B. The ensemble is used to mitigate single-model bias; we report the average of the two models' scores per dimension.

\paragraph{Prompt template.}
Both judges are given the user's history $\mathcal{H}_u$ (a sequence of images and captions, or reviews and product images, depending on the task) and a single generated image $\hat{I}_{u,r}$. They are asked to rate the generation on three dimensions, each on a $1$--$5$ Likert scale:

\begin{itemize}
\item \emph{Style}: how well the generation's visual style (color palette, composition, lighting, framing) matches the user's history.
\item \emph{Content}: how well the generation's subject matter and depicted activities match the user's history.
\item \emph{Overall}: an aggregate judgment of how plausibly the generation belongs to this particular user.
\end{itemize}

The full prompt template is reproduced in Appendix~\ref{app:prompts}. Scores are rounded to one decimal place and averaged across test instances and across the two judge models.

\subsection{Human Evaluation}
\label{app:human-study}

\paragraph{Scope and sampling.}
We report one annotator's complete evaluation of the original SDXL outputs, comprising 25 Personalized Scene Generation comparisons (\method{} versus PMG) and 25 Personalized Creative Generation comparisons (\method{} versus Pigeon), each involving a distinct test user. Later scene rerenderings and their judgments are not pooled with this round. Sampling used fixed seeds and available paired outputs, without inspecting output appearance or automatic scores. The scene sample (seed 20260918) contains five Home and Kitchen users and four each from Automotive, Clothing/Shoes/Jewelry, Electronics, Sports/Outdoors, and Tools/Home Improvement. The creative sample (seed 20260917) contains five users each from food, interior, pet, travel, and other, using the first held-out target for each user.

\paragraph{Annotation protocol.}
The interface displayed anonymized A/B candidates, the user history, and the generation condition. Scene cases also showed the reference product image and supplied product description. The annotator could inspect images at full size. Case order was randomized, with \method{} placed on the left in 13 scene and 12 creative comparisons and on the right in the remaining cases. The primary question asked which image better fit the person's history, with choices A, B, about equal, or insufficient evidence. A second question assessed product preservation for scenes and topic adherence for creative posts, allowing A, B, about equal, or neither satisfies it. All 50 responses were complete; none was excluded from the primary tabulation.

\paragraph{Displayed evidence.}
Scene histories contained eight or nine reviews with associated images. Creative cases showed the chronological union of the first five historical images used by \method{} and the last five used by Pigeon (nine or ten distinct images), excluding both held-out posts. Captions and generated prompts were not displayed for creative cases. Thus, the creative comparison evaluates the saved pipelines with their respective history subsets; it does not isolate method effects under identical conditioning.

\begin{table}[htbp]
\centering
\caption{\textbf{Human evaluation of the original SDXL outputs.} Entries are counts (percentages), with 25 pairs per task and question. The baseline is PMG for scenes and Pigeon for creative generation. ``Other'' denotes insufficient evidence for history fit and neither image satisfactory for product preservation or topic adherence.}
\label{tab:human-study}
\small
\setlength{\tabcolsep}{4pt}
\begin{tabular}{@{}llrrrr@{}}
\toprule
Task & Criterion & \method{} & Baseline & Equal & Other \\
\midrule
Scene & History fit & 19 (76\%) & 4 (16\%) & 0 (0\%) & 2 (8\%) \\
Scene & Product preservation & 8 (32\%) & 3 (12\%) & 9 (36\%) & 5 (20\%) \\
Creative & History fit & 16 (64\%) & 9 (36\%) & 0 (0\%) & 0 (0\%) \\
Creative & Topic adherence & 18 (72\%) & 5 (20\%) & 2 (8\%) & 0 (0\%) \\
\bottomrule
\end{tabular}
\end{table}

\paragraph{Results and interpretation.}
Table~\ref{tab:human-study} reports every response category, retaining all 25 cases per task in each percentage denominator. History-fit choices favor \method{} in this sample, while scene product preservation is less decisive: nine pairs are judged equal and five unsatisfactory for both methods. Creative topic-adherence choices favor \method{} in 18 cases, Pigeon in five, with two ties. This single-annotator evaluation measures an outside observer's assessment of fit to the available history. It does not establish inter-annotator reliability, population-level preference, or statistical significance.

\paragraph{Scene input uncertainty.}
One scene account was subsequently flagged for target-review leakage in a separate rerendering run. Its original SDXL generation inputs are unavailable for verification, so contamination of the original outputs is unresolved. As a sensitivity check, excluding this account from the original responses yields 18/24 history-fit choices for \method{} (75\%), four for PMG, and two insufficient-evidence responses. Product-preservation counts become eight for \method{}, three for PMG, nine ties, and four neither-satisfactory responses. The primary table retains the complete original sample and does not substitute rerendered outputs.

\section{Additional Results}

\subsection{Ablation Studies}
\label{app:ablation}
\begin{table*}[t]
    \small
    \centering
    \caption{Ablation Study for Personalized Scene Generation.
    \textbf{Bold} marks the best per column.}
    \label{tab:task1_ablation_sft}
    \small
    \setlength{\tabcolsep}{4pt}
    \begin{tabular}{l cc c ccc}
    \toprule
    & \multicolumn{2}{c}{Retrieval} & Image Quality & \multicolumn{3}{c}{MLLM Judge ($1$--$5$)} \\
    \cmidrule(lr){2-3} \cmidrule(lr){4-4} \cmidrule(lr){5-7}
    Method & H@5\,$\uparrow$ & MRR\,$\uparrow$ & Aes\,$\uparrow$ & Cont.\,$\uparrow$ &
  Overall\,$\uparrow$ & Sty.\,$\uparrow$ \\
    \midrule
    \method{}-Reflection     & \textbf{0.2336} & 0.1583 & 5.32 & 4.14 & \textbf{3.99} & 2.74 \\
    \method{} (ours)  & 0.2297 & \textbf{0.1639} & \textbf{5.33} & \textbf{4.36} & 3.92 &
  \textbf{2.88} \\
    \bottomrule
    \end{tabular}
  \end{table*}
\begin{table*}[t]
  \centering
  \caption{Ablation Study for Personalized Creative Generation.
  \textbf{Bold} marks the better of the two per column.}
  \label{tab:task2_sft_vs_rl}
  \footnotesize
  \setlength{\tabcolsep}{2.5pt}
  \begin{tabular}{l cccc cc ccc}
  \toprule
  & \multicolumn{4}{c}{Target-based Image Metrics} & \multicolumn{2}{c}{Contrastive R@1} &
  \multicolumn{3}{c}{MLLM Judge ($1$--$5$)} \\
  \cmidrule(lr){2-5} \cmidrule(lr){6-7} \cmidrule(lr){8-10}
  Method & CIS\,$\uparrow$ & DIS\,$\uparrow$ & LPIPS\,$\downarrow$ & MS-SSIM\,$\uparrow$ &
  Inter-cat.\,$\uparrow$ & Intra-cat.\,$\uparrow$ & Sty.\,$\uparrow$ & Cont.\,$\uparrow$ &
  Overall\,$\uparrow$ \\
  \midrule
  PEARL-Reflection                                                              & 0.635 & 0.285 &
  \textbf{0.732} & \textbf{0.057} & 0.838 & 0.701 & 3.529 & \textbf{3.331} & 3.516 \\
  \method{} (ours)                                                       & \textbf{0.645} &
  \textbf{0.289} & 0.735 & 0.055 & \textbf{0.876} & 0.701 & \textbf{3.554} & 3.277 & \textbf{3.541}
  \\
  \bottomrule
  \end{tabular}
  \end{table*}

We provide the full ablation study results in Table~\ref{tab:task1_ablation_sft} and Table~\ref{tab:task2_sft_vs_rl} for Personalized Scene Generation and Personalized Creative Generation, respectively. Across both tasks, \method{} consistently outperforms \method{}-SFT on the MLLM judge metrics, with the largest margins appearing on style and content scores. This pattern indicates that the alternating-policy DPO stage primarily contributes to fine-grained personalization details that the reflector identifies and corrects when conditioning on rendered evidence, rather than to coarse changes in scene layout or subject matter that the planner alone can already produce. The retrieval and image-quality metrics show smaller and less consistent differences between the two variants, which is expected: surface-level retrieval encoders pool over global visual features and cannot fully resolve the fine-grained stylistic adjustments where the reflector contributes most. Notably, the marginal drop on the overall MLLM judge score (under $1\%$) suggests that the reflector concentrates its gains on fine-grained personalization details rather than on drastic adjustments to the rendered image. This is consistent with the findings of IRG~\citep{huang2025interleavingreasoningbettertexttoimage}, which similarly observed that interleaved reasoning yields its largest gains on fine-grained fidelity rather than on coarse semantic alignment.

\subsection{Qualitative Results}
\label{app:qualitative}

\begin{figure*}[!htbp]
\centering

\begin{minipage}{\textwidth}
\centering
\textbf{User 1} \\[3pt]
\setlength{\tabcolsep}{2pt}
\begin{tabular}{@{}cccccc@{}}
\includegraphics[width=0.14\textwidth]{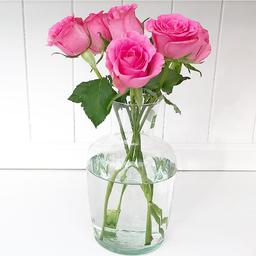} &
\includegraphics[width=0.14\textwidth]{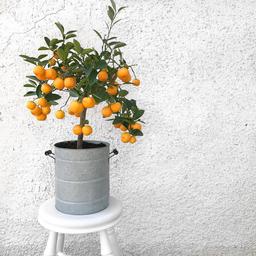} &
\includegraphics[width=0.14\textwidth]{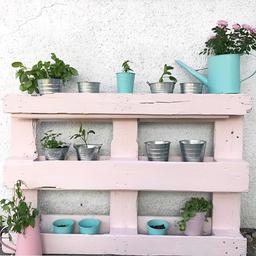} &
\includegraphics[width=0.14\textwidth]{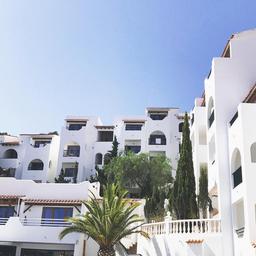} &
\includegraphics[width=0.14\textwidth]{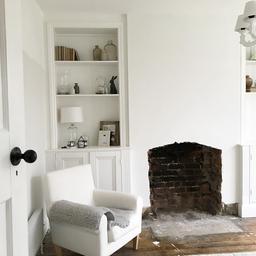} &
\includegraphics[width=0.14\textwidth]{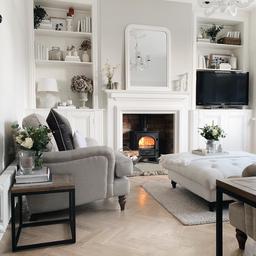} \\
\multicolumn{5}{c}{\footnotesize User History} & \footnotesize Held-out Target
\end{tabular}\\[3pt]
\begin{tabular}{@{}ccccc@{}}
\includegraphics[width=0.14\textwidth]{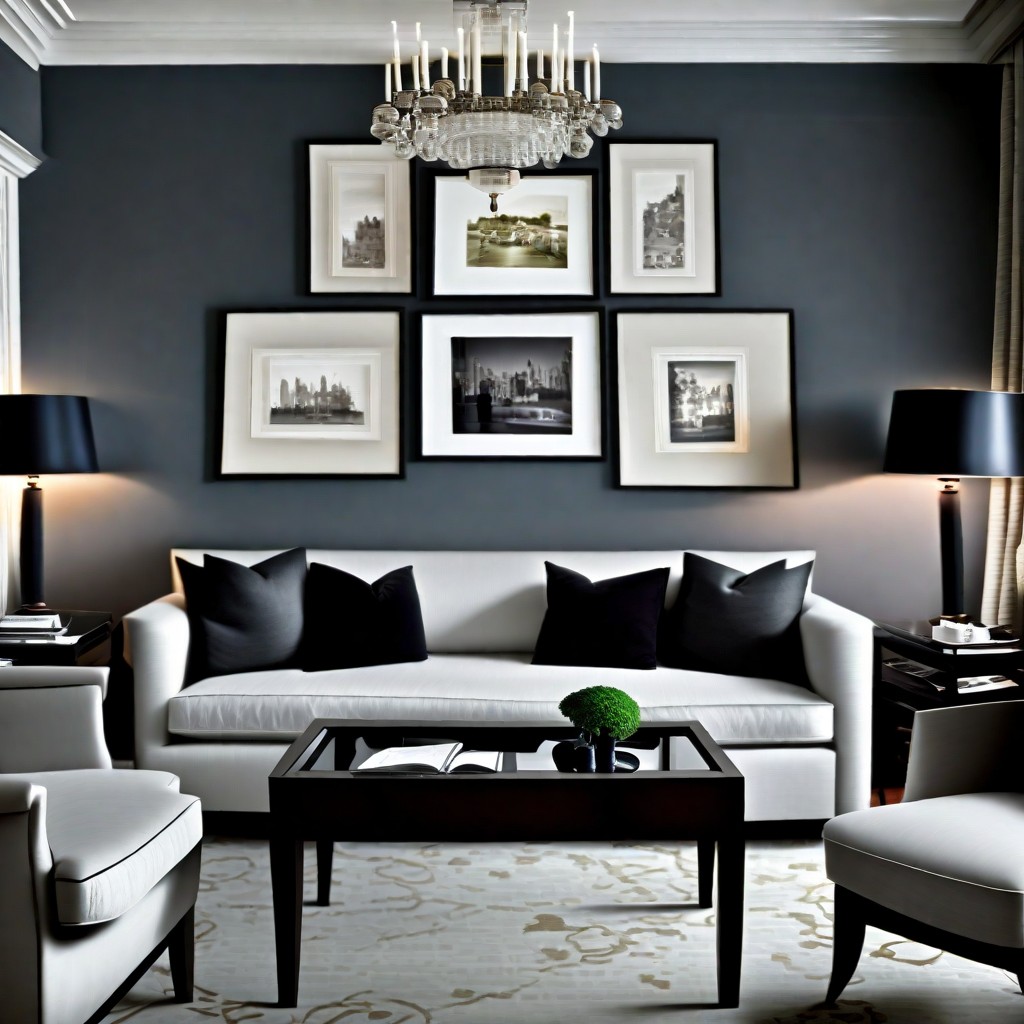} &
\includegraphics[width=0.14\textwidth]{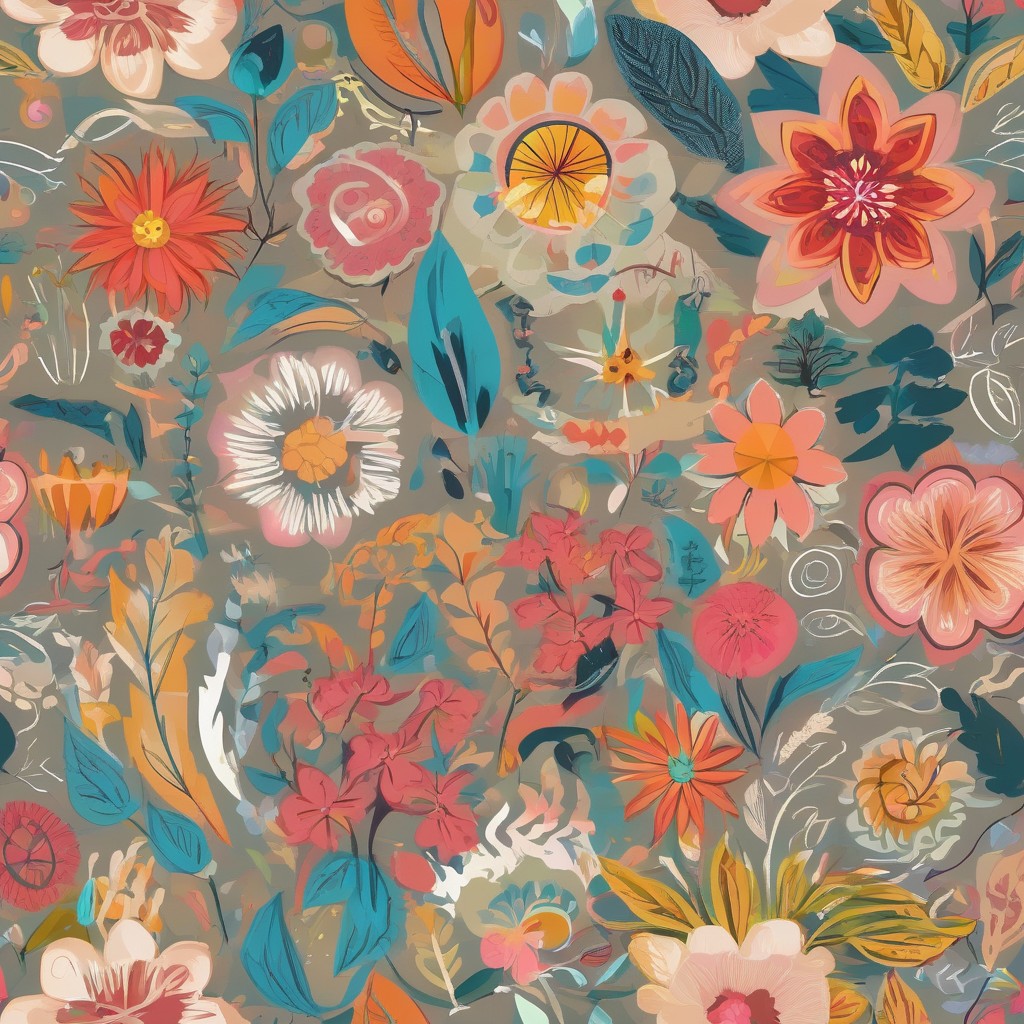} &
\includegraphics[width=0.14\textwidth]{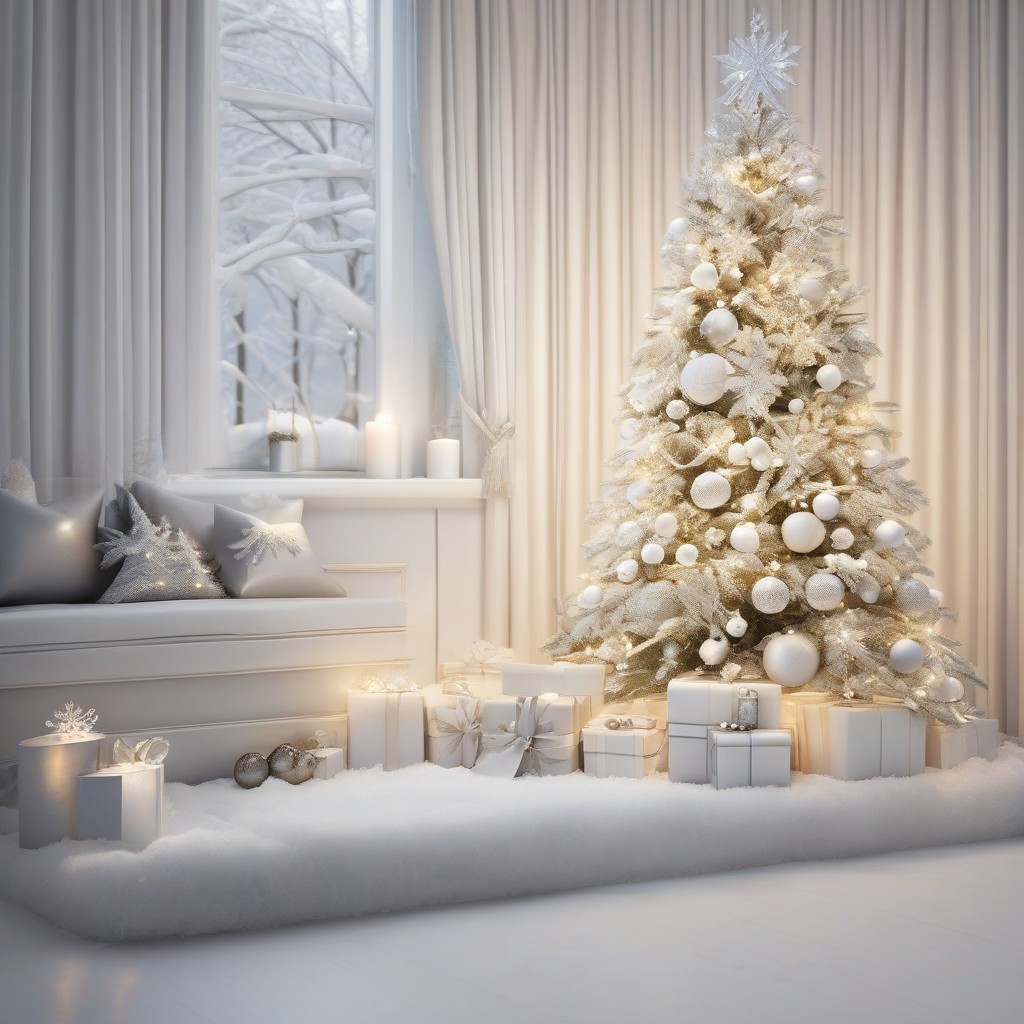} &
\includegraphics[width=0.14\textwidth]{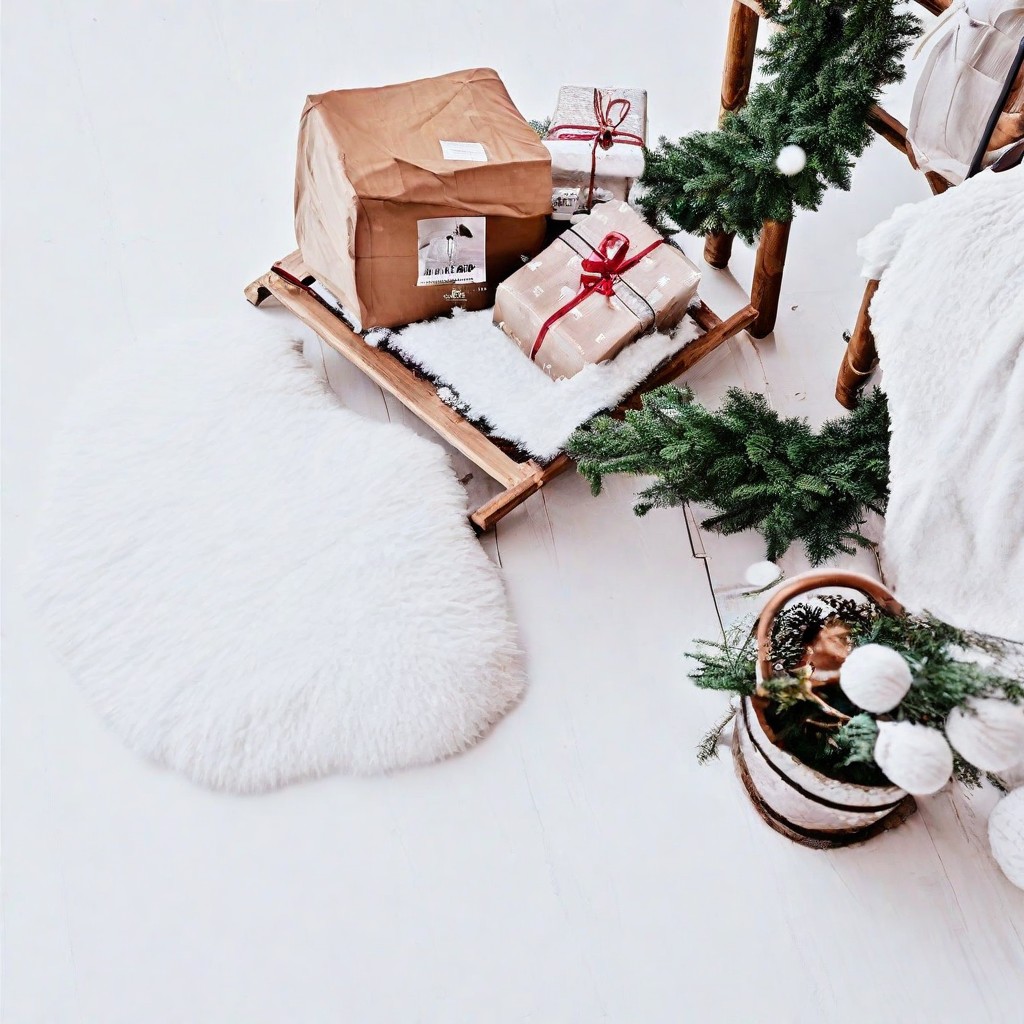} &
\includegraphics[width=0.14\textwidth]{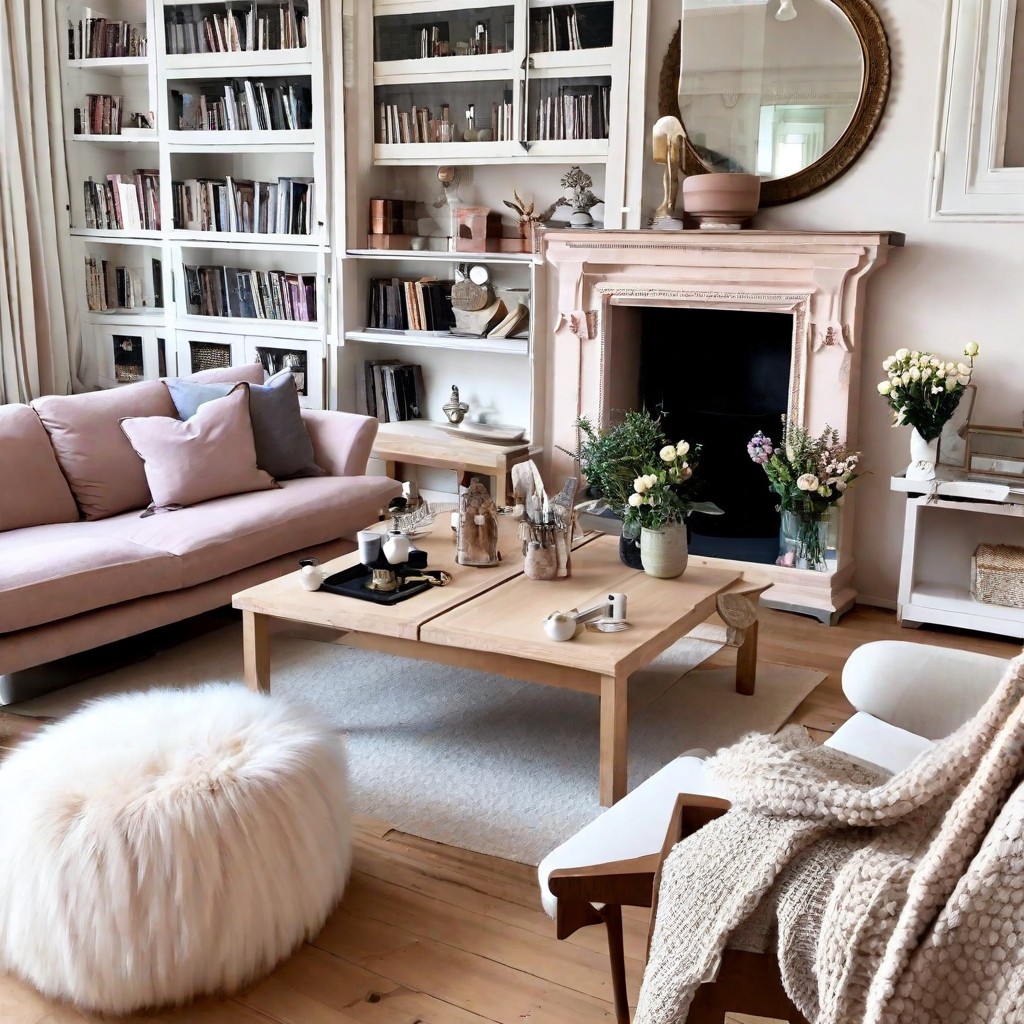} \\
\footnotesize LaVIT & \footnotesize LLaVA & \footnotesize PMG & \footnotesize Pigeon & \footnotesize \textbf{\method{}}
\end{tabular}
\end{minipage}

\vspace{12pt}

\begin{minipage}{\textwidth}
\centering
\textbf{User 2}\\[3pt]
\setlength{\tabcolsep}{2pt}
\begin{tabular}{@{}cccccc@{}}
\includegraphics[width=0.14\textwidth]{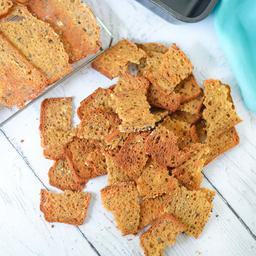} &
\includegraphics[width=0.14\textwidth]{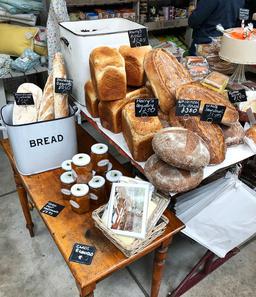} &
\includegraphics[width=0.14\textwidth]{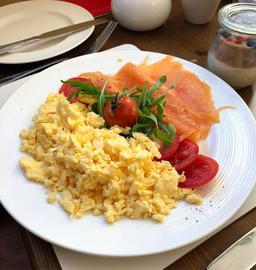} &
\includegraphics[width=0.14\textwidth]{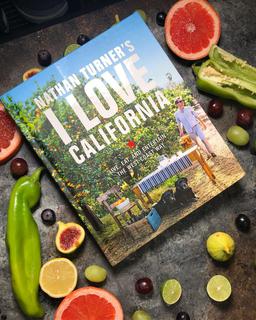} &
\includegraphics[width=0.14\textwidth]{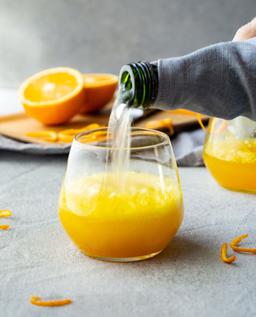} &
\includegraphics[width=0.14\textwidth]{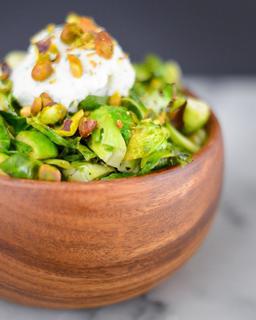} \\
\multicolumn{5}{c}{\footnotesize User History} & \footnotesize Held-out Target
\end{tabular}\\[3pt]
\begin{tabular}{@{}ccccc@{}}
\includegraphics[width=0.14\textwidth]{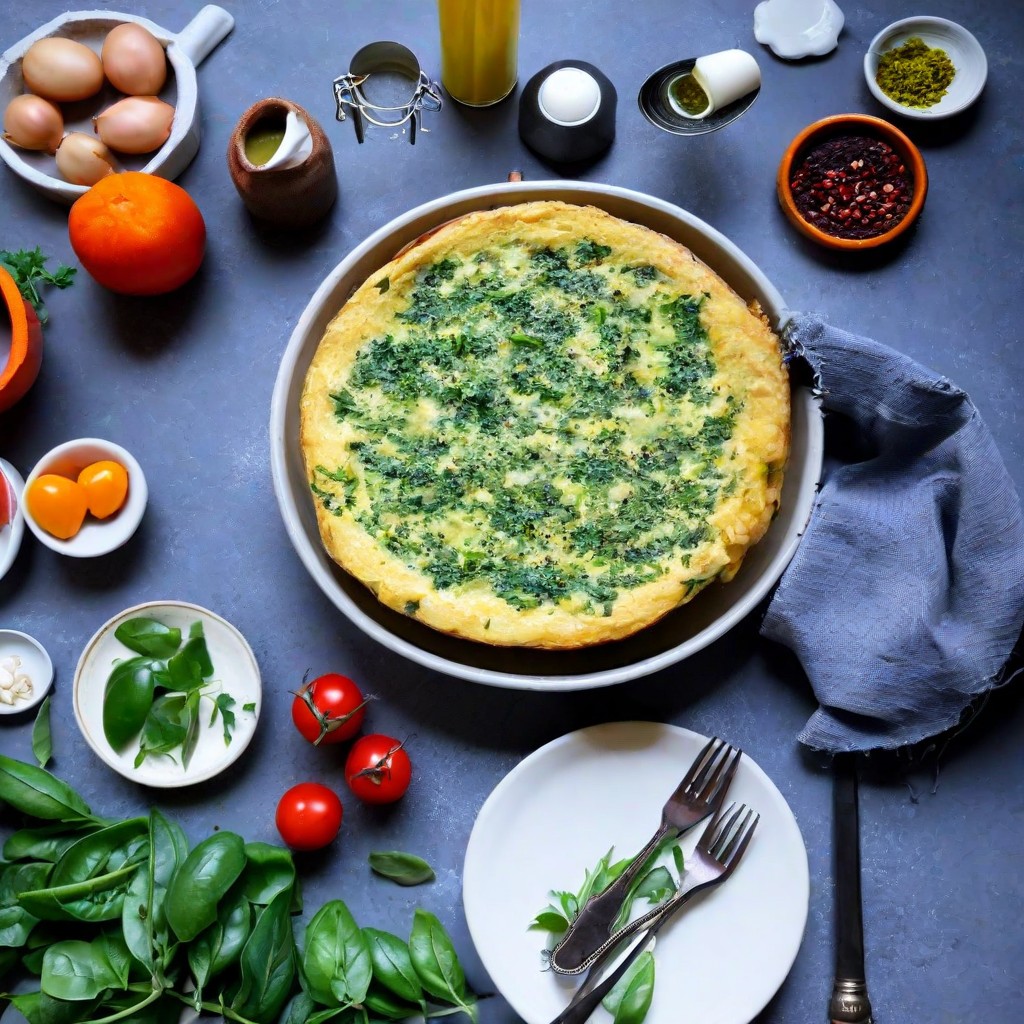} &
\includegraphics[width=0.14\textwidth]{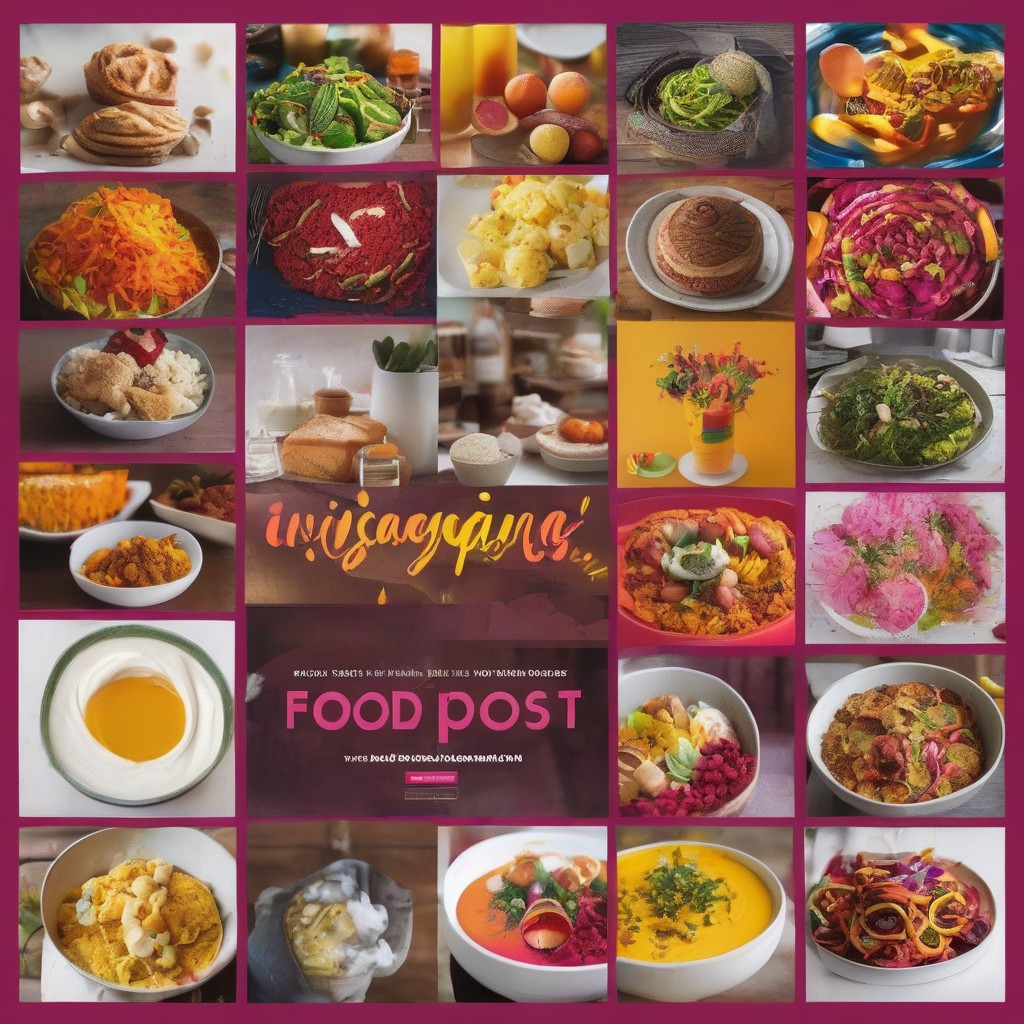} &
\includegraphics[width=0.14\textwidth]{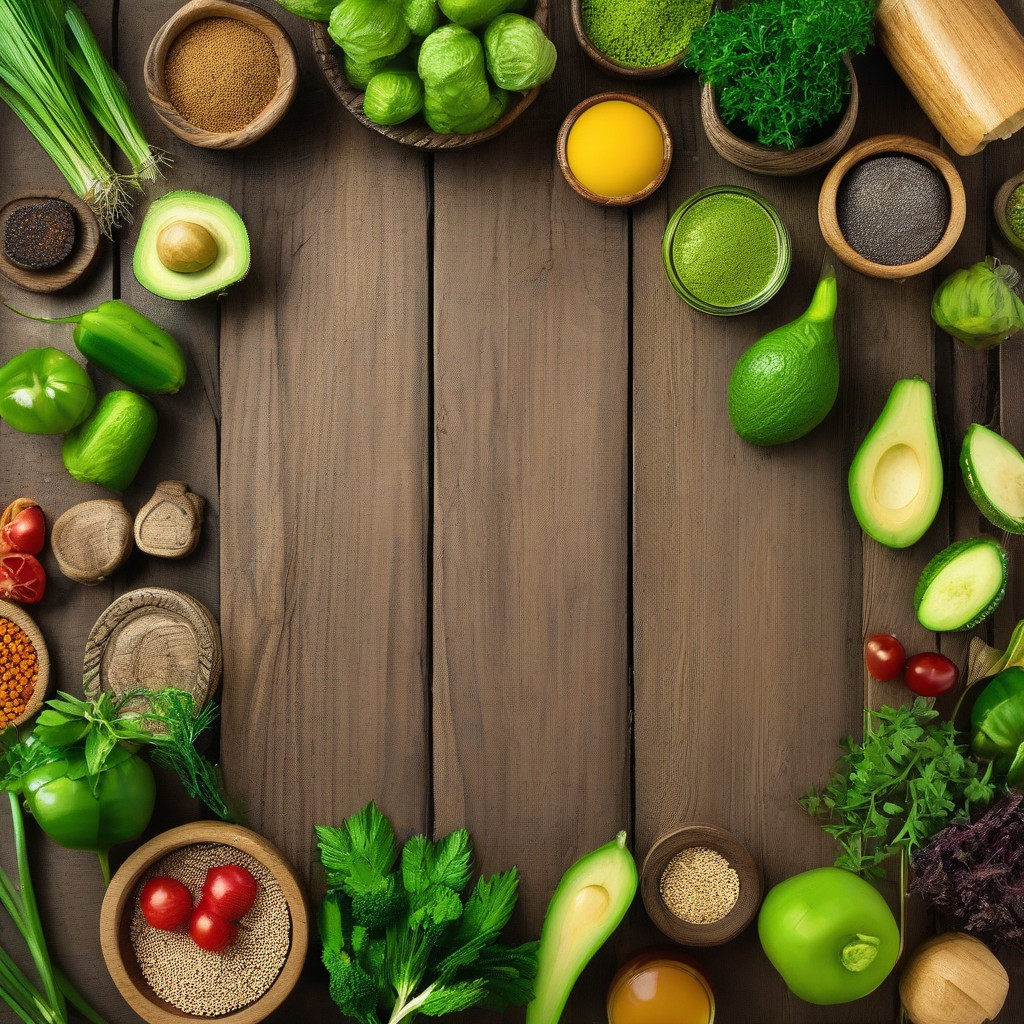} &
\includegraphics[width=0.14\textwidth]{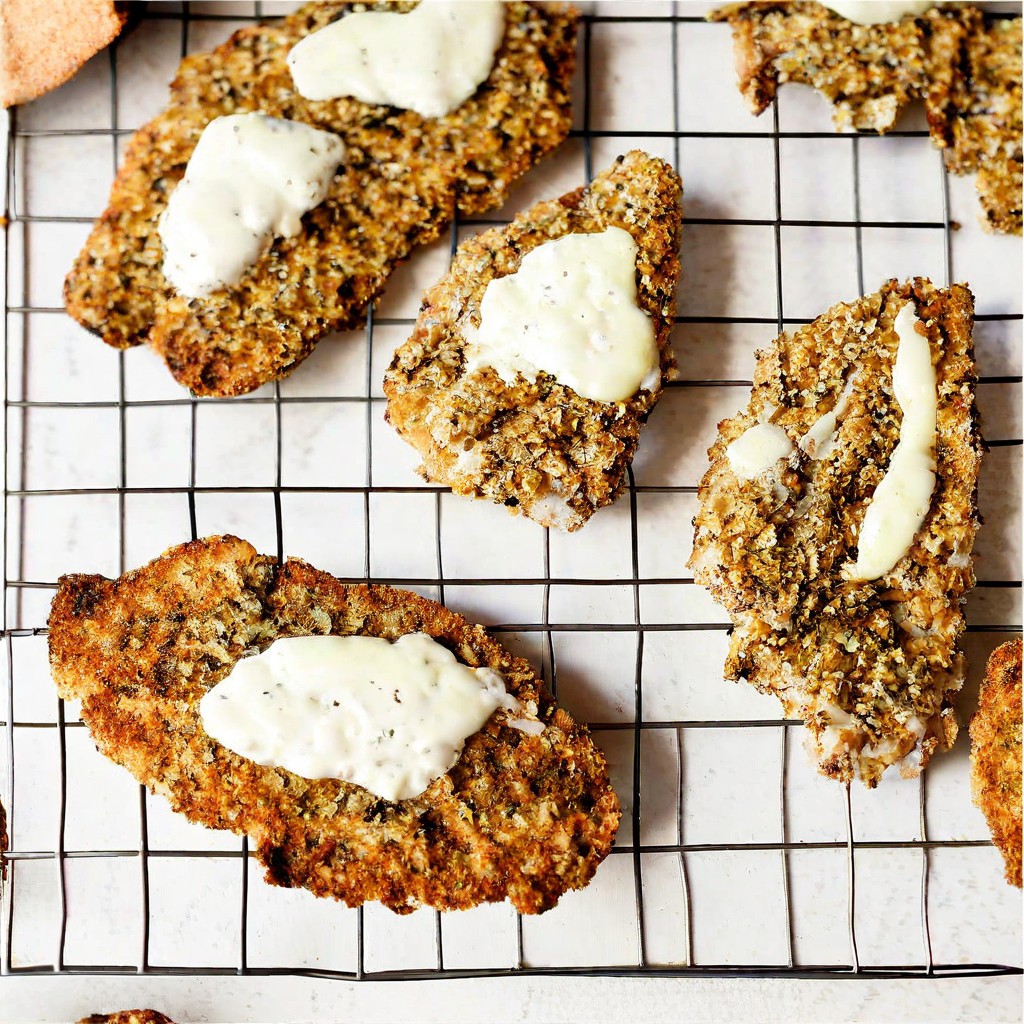} &
\includegraphics[width=0.14\textwidth]{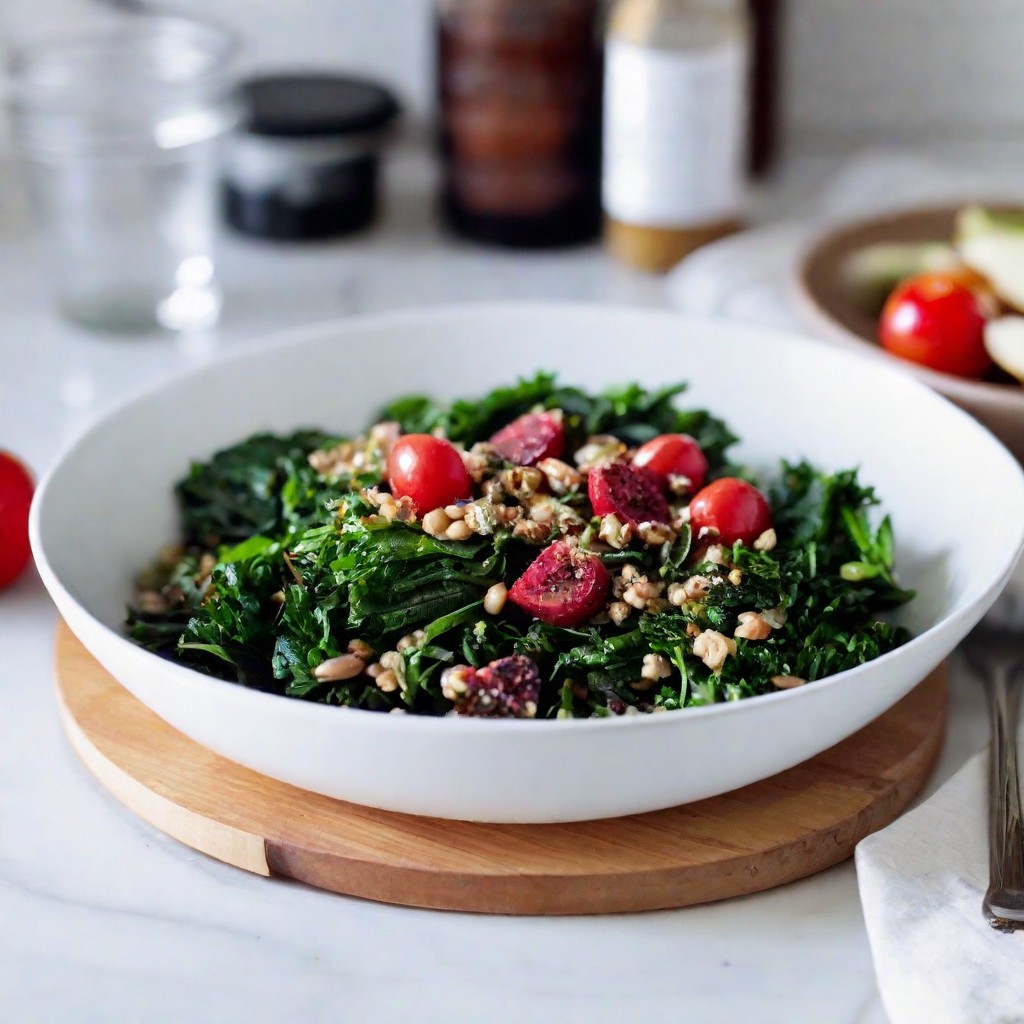} \\
\footnotesize LaVIT & \footnotesize LLaVA & \footnotesize PMG & \footnotesize Pigeon & \footnotesize \textbf{\method{}}
\end{tabular}
\end{minipage}

\vspace{12pt}

\begin{minipage}{\textwidth}
\centering
\textbf{User 3} \\[3pt]
\setlength{\tabcolsep}{2pt}
\begin{tabular}{@{}cccccc@{}}

\includegraphics[width=0.14\textwidth]{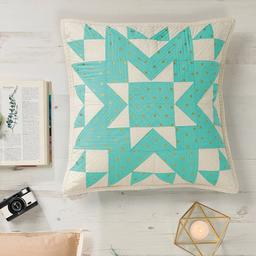} &
\includegraphics[width=0.14\textwidth]{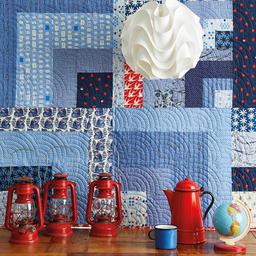} &
\includegraphics[width=0.14\textwidth]{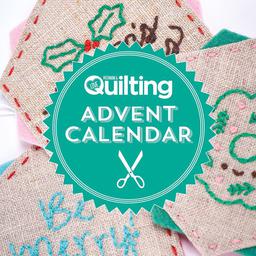} &
\includegraphics[width=0.14\textwidth]{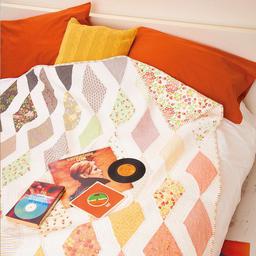} &
\includegraphics[width=0.14\textwidth]{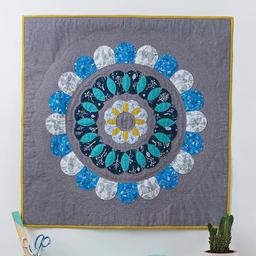} &
\includegraphics[width=0.14\textwidth]{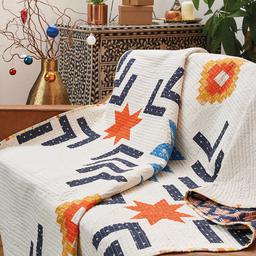} \\
\multicolumn{5}{c}{\footnotesize User History} & \footnotesize Held-out Target
\end{tabular}\\[3pt]
\begin{tabular}{@{}ccccc@{}}
\includegraphics[width=0.14\textwidth]{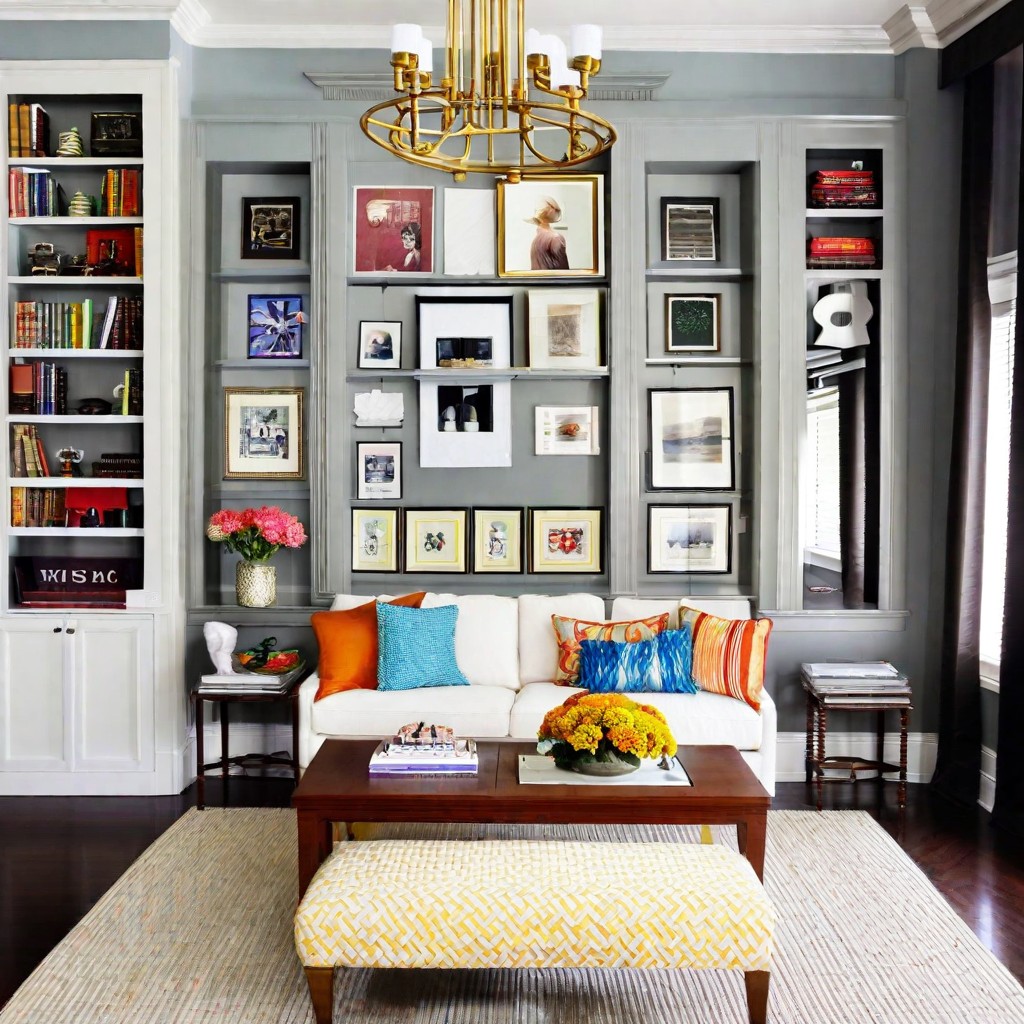} &
\includegraphics[width=0.14\textwidth]{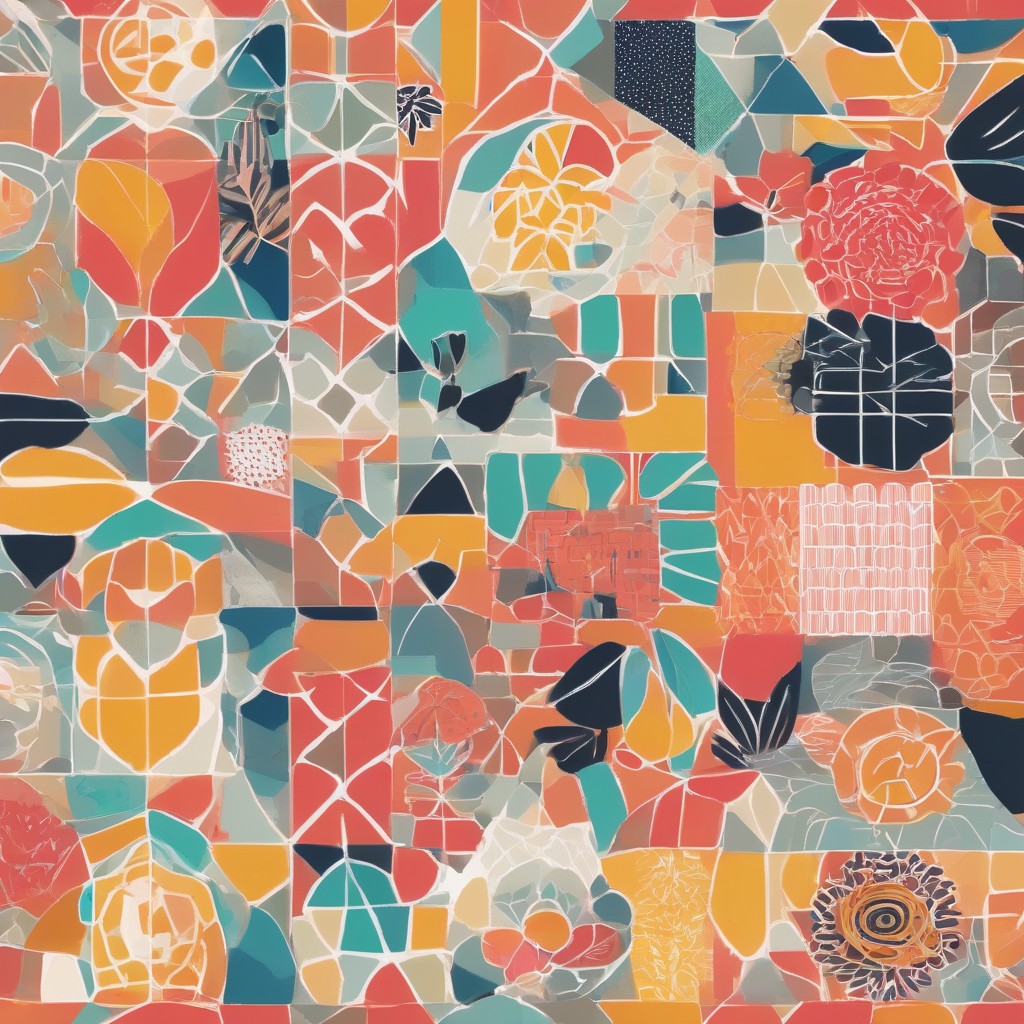} &
\includegraphics[width=0.14\textwidth]{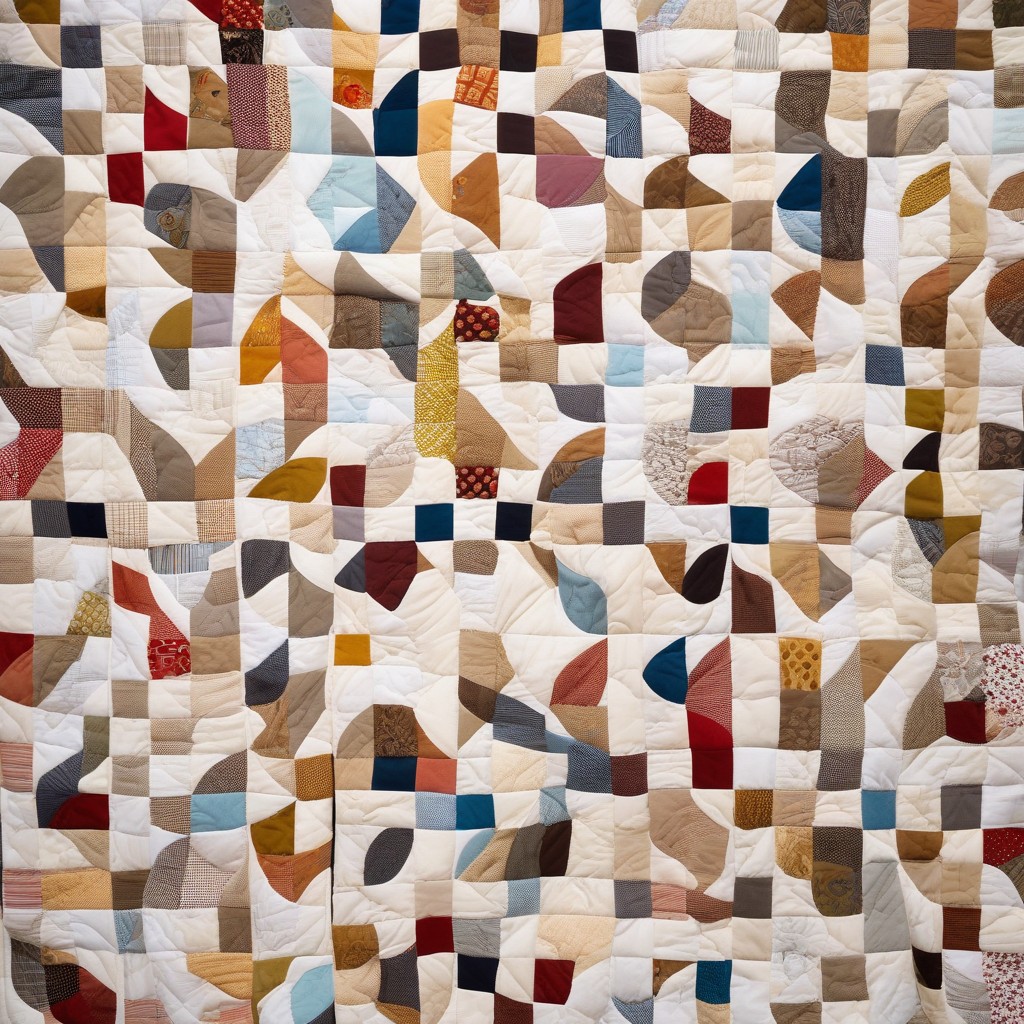} &
\includegraphics[width=0.14\textwidth]{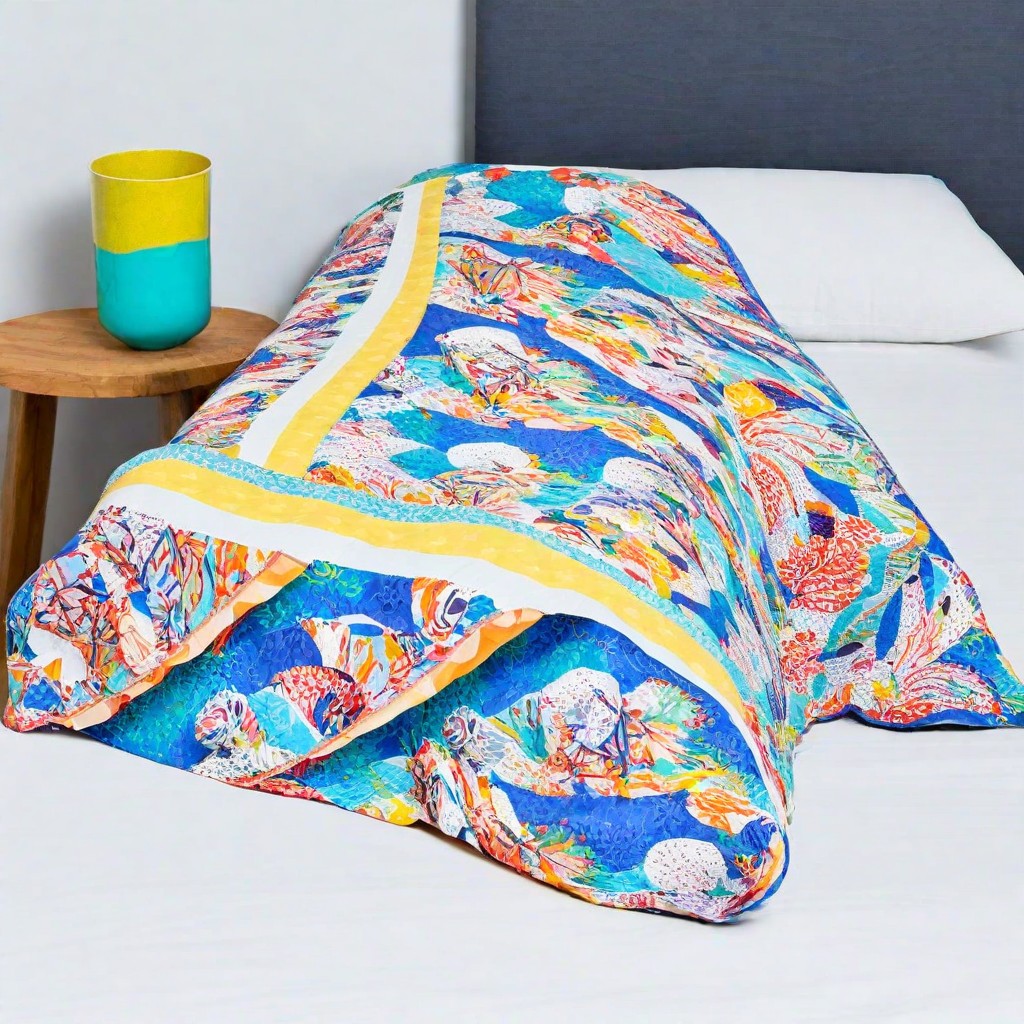} &
\includegraphics[width=0.14\textwidth]{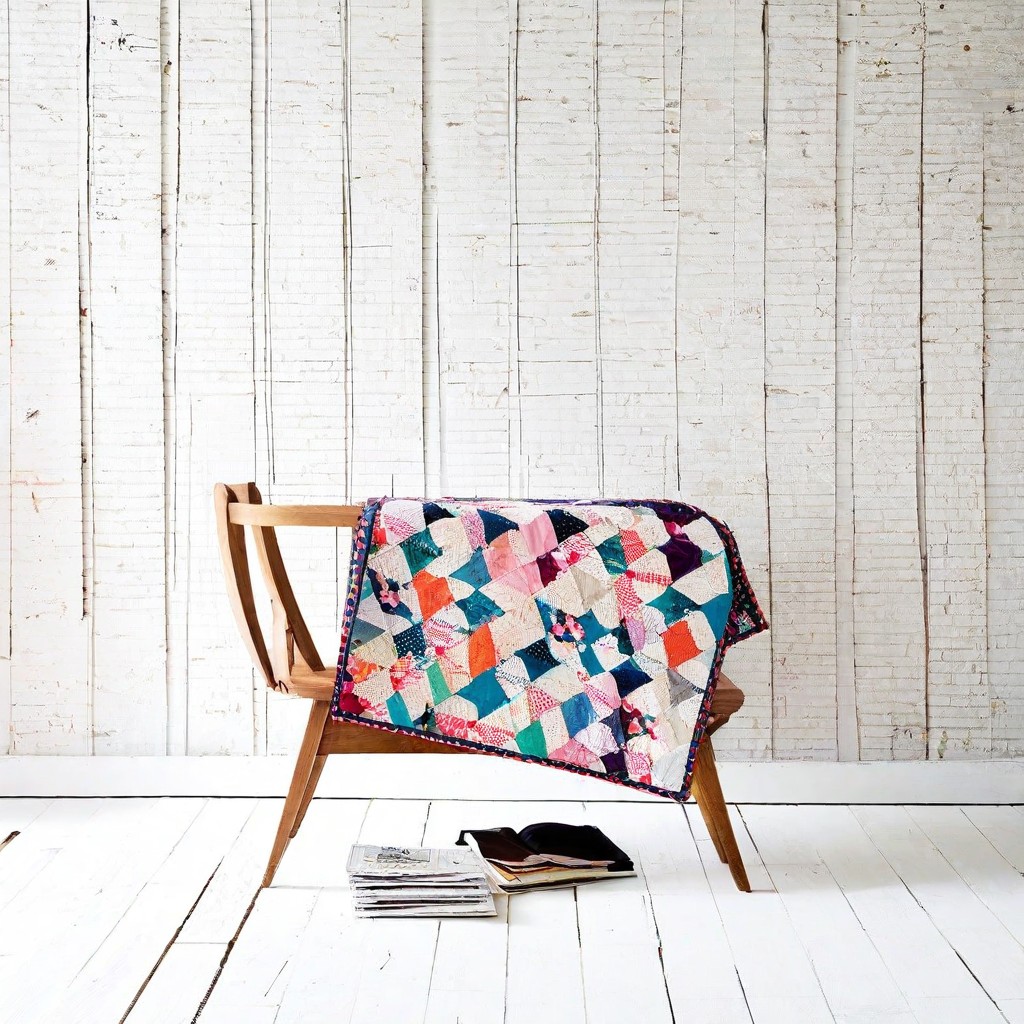} \\
\footnotesize LaVIT & \footnotesize LLaVA & \footnotesize PMG & \footnotesize Pigeon & \footnotesize \textbf{\method{}}
\end{tabular}
\end{minipage}

\caption{\textbf{Additional qualitative results on Personalized Creative Generation.} For each of three users, the top row shows five historical posts alongside a held-out target post (rightmost), and the bottom row shows generations from LaVIT, LLaVA, PMG, Pigeon, and \method{}. The examples illustrate recurring choices in color, framing, and scene presentation across interiors, food, and quilting content.}
\label{fig:app-task2-qualitative}
\end{figure*}

We provide additional qualitative results on Personalized Creative Generation in Figure~\ref{fig:app-task2-qualitative}. The three users illustrate recurring posting preferences in interiors, food, and quilting. For User 1, the displayed history contains light backgrounds, plants, and a fireplace scene. \method{} retains the light-colored palette and includes a fireplace in a furnished living room, while LaVIT depicts a darker interior, LLaVA produces a floral pattern, and PMG and Pigeon emphasize holiday decorations.

For User 2, the history includes close views of food and drinks presented on tables or counters, with natural-looking light and visible ingredient textures. \method{} produces a salad in a white bowl on a kitchen counter, retaining this photographic presentation. LLaVA instead produces a collage with text, and PMG arranges ingredients around an empty central area; LaVIT and Pigeon also retain plausible food-photography cues.

For User 3, the history repeatedly presents colorful geometric textiles in domestic settings. \method{} depicts a patchwork quilt draped over a chair against a light wooden wall. LLaVA and PMG emphasize flat patterns, while LaVIT depicts a wider furnished room without a prominent quilt; Pigeon also captures a quilt displayed on furniture. These examples illustrate alignment with recurring choices in color, framing, and setting across different subjects, without establishing exact reproduction of an individual subject.

\newpage
\section{Prompts}
\label{app:prompts}

\begin{tcolorbox}[title=Task 1 Planning, colframe=gray]
\scriptsize   
\begin{verbatim}
System Prompt: You write a single-trait background prompt for an inpaint pipeline. 
The target product is pasted into the final image as-is; you describe only the scene 
AROUND the product, grounded in user identity element.

Review history:
  {reviews_block}

Target product:
  {product_description}

Image 1 is the product catalog photo (shown for context only; it will be
pasted into the output). Select one identity element from the reviews and
write a background-only scene prompt grounded in it.

Output in this EXACT 3-section format (tags literal, in order):

<chosen>
[one identity element, copied verbatim from the reviews. A brand, hobby, or
context.]
</chosen>
<why>
[One sentence: why this element fits this product better than others.]
</why>
<prompt>
  [AT MOST 40 words describing the BACKGROUND scene around the product.
  Describe ONLY setting, furniture/props, lighting, and atmosphere.
  Must include the chosen element as a concrete scene noun.
  Do NOT mention the product, its name, or anything that could produce a
  duplicate product in the scene.]
</prompt>
\end{verbatim}
   
\end{tcolorbox}

\begin{tcolorbox}[title=Task 1 Reflection, colframe=gray]
\scriptsize   
\begin{verbatim}
System Prompt: You reflect on an AI-generated BACKGROUND scene for an inpaint pipeline. 
The target product is pasted into the image as-is; you critique and refine ONLY the 
surrounding scene, grounded in one user identity element.

Review history:
  {reviews_block}

Target product:
  {product_description}

The first image is the product catalog photo (context only). The second
image is a first-attempt BACKGROUND scene around the product. Critique the background 
and write a refined BACKGROUND-only prompt.

Output in this EXACT 4-section format (tags literal, in order):

<analysis>
  [What the background currently shows and which aspects of the user's
  identity are missing / misrepresented. Focus on the scene, not the product.]
</analysis>
<improvements>
  [What a refined BACKGROUND should include. 2-3 concrete scene elements
  grounded in user identity. Do NOT describe the product.]
</improvements>
<modifications>
  [Step-by-step concrete bg changes. Prefer ADDING over REPLACING.]
</modifications>
<prompt>
  [Refined BACKGROUND description --- AT MOST 40 words. Setting/props/lighting
  grounded in user identity. Do NOT mention the product or anything that
  could create a duplicate product.]
</prompt>
\end{verbatim}
   
\end{tcolorbox}

\newpage

\begin{tcolorbox}[title=Task 2 Planning, colframe=gray]
\scriptsize   
\begin{verbatim}
System Prompt: You are a visual style analyst helping create personalized image 
generation prompts. You will see a social media user's posting history. Your 
job is to analyze their visual style, then write a detailed scene prompt that would 
produce a new image matching their aesthetic. Captions may be in any language. Always 
respond in English.

[5 history images]

Account: @{username} | Category: {category} | Followers: {followers}

The images above are this user's recent posts (history).

Analyze the user's visual style, then write a detailed scene prompt for a new {topic} 
image that would match their aesthetic.

Output in this exact format (use the tags literally):

<think>
  [2-4 sentences analyzing who this user is, their visual style patterns ---
  color palette, lighting, composition, textures, mood, recurring elements]
</think>
<prompt>
  [A detailed, specific scene prompt for a new image that fits this user's
  visual style. Include specifics about colors, lighting, composition, objects,
  textures, atmosphere. Write it as a standalone image generation prompt.]
</prompt>
\end{verbatim}
   
\end{tcolorbox}

\begin{tcolorbox}[title=Task 2 Reflection, colframe=gray]
\scriptsize   
\begin{verbatim}
You are a visual style reflection reasoner. Given a user's posting history and 
a first-attempt generated image, critique the attempt and produce a refined prompt
that better matches the user's aesthetic.

[5 history images]
[1 initial image generated by Stage 1]

Account: @{username} | Category: {category}

Images 1-{n_history}: This user's recent posting history.
Image {n_history_plus_1}: An AI-generated image attempt for this user.

The attempt was generated from this prompt:
"{stage1_prompt}"

Compare the attempt with the user's posting history. Identify STYLE differences
(colors, lighting, composition, mood) and produce a refined prompt.

IMPORTANT: Keep the SAME SUBJECT as the original prompt. Only modify style elements.

Your response must follow this exact structure:

<think>
[What the attempt got right and what STYLE elements it misses vs the user's history.]

### Detailed Explanation of Required Improvements:
[Style/composition/lighting/color issues only. Do NOT suggest changing the subject.]

### Step-by-Step Modification Guidance:
[2-4 steps that modify style while keeping the same subject.]

### Final Comprehensive Prompt for the Improved Image:
[A refined version of the ORIGINAL prompt. Same subject, improved style descriptors.]
</think>
\end{verbatim}
   
\end{tcolorbox}

\begin{tcolorbox}[title=Task 1 MLLM-as-Judge, colframe=gray]
\scriptsize   
\begin{verbatim}
  You are evaluating an AI-generated image meant to be a personalized
  product scene for a specific Amazon user. The user is described by their
  review history (text), and the image should depict the target product
  naturally placed within a scene that fits the user's life.

  User's review history:
  {reviews_block}

  Target product: {product_description}

  The image above is the AI-generated personalized scene. Rate it on three
  aspects (integer 1-5, where 5 = excellent match):

  - style: Does the scene's visual style (lighting, palette, composition,
    mood) match the aesthetic implied by this user's reviews and lifestyle?
  - content: Are the scene's setting / props / activities
    identifiably grounded in this user's identity (hobbies, recurring
    interests, household)? A generic "lifestyle stock photo" should score
    low; a scene that obviously belongs to *this* user should score high.
  - overall: Overall quality as a personalized product scene for this user
    (combining visual quality, product integration, and identity fit).

  Output ONLY:
  style: <1-5>
  personalization: <1-5>
  overall: <1-5>
\end{verbatim}
   
\end{tcolorbox}

\begin{tcolorbox}[title=Task 2 MLLM-as-Judge, colframe=gray]
\scriptsize   
\begin{verbatim}
  You are evaluating an AI-generated image meant to replicate a specific
  Instagram user's style.

  Image 1 is a real post by the user (ground truth).
  Image 2 is an AI-generated image intended to match that user's style.

  Rate Image 2 on three aspects (integer 1-5, where 5 = excellent match):

  - style_match: Does Image 2 match Image 1's visual style (lighting, colors,
    composition, mood)?
  - content_match: Is Image 2's subject/scene appropriate given Image 1?
  - overall: Overall quality as a personalized post matching this user.
\end{verbatim}
   
\end{tcolorbox}

\section{Notations}
\label{app:notation}
\begin{table}[h]
\centering
\small
\caption{Notation used throughout the paper.}
\label{tab:notation}
\renewcommand{\arraystretch}{1.15}
\begin{tabular}{@{}p{0.18\linewidth}p{0.78\linewidth}@{}}
\toprule
\textbf{Symbol} & \textbf{Description} \\
\midrule
\multicolumn{2}{@{}l@{}}{\textit{Problem Setup}} \\
\midrule
$u$ & A user. \\
$\mathcal{H}_u$ & User $u$'s multimodal history; a sequence of prior activities. \\
$h^{(u)}_i$ & The $i$-th history entry for user $u$ (e.g., a review with product image, or a post with caption). \\
$N_u$ & Number of history entries for user $u$. \\
$T$ & A personalized generation task. \\
$\mathcal{C}_T$ & Condition space associated with task $T$. \\
$c \in \mathcal{C}_T$ & Generation condition; specifies what the generated image should depict or preserve. \\
$\hat{I}_{u,c}$ & Personalized image generated for user $u$ under condition $c$. \\
$f^T_\theta$ & Task-specific parameterized model mapping $(\mathcal{H}_u, c)$ to $\hat{I}_{u,c}$. \\
\midrule
\multicolumn{2}{@{}l@{}}{\textit{Task Instantiations}} \\
\midrule
$(\ell_i, o_i)$ & Personalized Scene Generation history entry: review $\ell_i$ paired with product image $o_i$. \\
$o^\star$ & Target product image (Personalized Scene Generation condition). \\
$(z_i, m_i)$ & Personalized Creative Generation history entry: post image $z_i$ paired with caption metadata $m_i$. \\
$t^\star$ & Target topic or content specification (Personalized Creative Generation condition). \\
\midrule
\multicolumn{2}{@{}l@{}}{\textit{\method{} Framework}} \\
\midrule
$\pi_\phi$ & Planner policy with parameters $\phi$. \\
$\pi_\psi$ & Reflector policy with parameters $\psi$. \\
$G$ & Frozen text-to-image generator (renderer). \\
$r_1$ & Planner reasoning trace summarizing implications of $\mathcal{H}_u$. \\
$s_1$ & Initial scene description emitted by the planner. \\
$I_1 = G(s_1, c)$ & Initial rendering produced by the renderer from $s_1$. \\
$r_2$ & Reflector structured analysis comparing $I_1$ against $\mathcal{H}_u$. \\
$s_2$ & Revised scene description emitted by the reflector. \\
$\hat{I} = G(s_2, c)$ & Final rendered image after reflection. \\
\midrule
\multicolumn{2}{@{}l@{}}{\textit{Training}} \\
\midrule
$\mathcal{T}$ & Teacher multimodal model used for silver trajectory distillation. \\
$\mathcal{I}^\star_u$ & Ground-truth identity signal accessible to the teacher (held-out target image or extracted identity elements). \\
$(r_1^\star, s_1^\star)$ & Silver planner trajectory produced by $\mathcal{T}$. \\
$(r_2^\star, s_2^\star)$ & Silver reflector trajectory produced by $\mathcal{T}$. \\
$\tilde{I}_1$ & Pseudo-initial image rendered from $s_1^\star$ during silver supervision. \\
$\mathcal{L}_{\text{SFT}}$ & Cross-entropy loss for Stage 1 supervised fine-tuning. \\
$\mathcal{L}_{\text{DPO}}$ & Direct Preference Optimization loss for Stage 2. \\
$\pi_\phi^{\text{ref}}, \pi_\psi^{\text{ref}}$ & Reference (warm-started) policies frozen during DPO. \\
$\beta$ & DPO temperature. \\
$K$ & Number of candidate trajectories sampled per preference pair. \\
$R(\hat{I}, \mathcal{H}_u)$ & Task-aligned retrieval reward used to score candidates. \\
$(\cdot)^w, (\cdot)^l$ & Winning and losing candidates in a DPO preference pair. \\
\bottomrule
\end{tabular}
\end{table}
Table~\ref{tab:notation} provides a summary and serves as a reference guide for notations used throughout the paper. 

\section{Applications and Use Cases}
\label{sec:applications}
The problem formulation, benchmark, and evaluation protocol introduced in this work are designed around two concrete instantiations, Personalized Scene Generation and Personalized Creative Generation, but the underlying task structure is general: given a user's naturally occurring multimodal history and a target specification, generate a novel image that reflects that user's identity. This generality makes the benchmark applicable across a broad range of settings beyond the e-commerce and social media domains studied in the main paper. On the generation side, any task that requires situating a given visual element in a user-appropriate context can be cast as a variant of Personalized Scene Generation, including personalized advertising (Section~\ref{sec:advertising}), lifestyle-aware product staging (Section~\ref{sec:tryon}), and object-to-scene synthesis as an inversion of the conventional image editing paradigm (Section~\ref{sec:inverse-editing}). On the creative side, any task that requires extending a user's established visual identity to a new subject can be cast as a variant of Personalized Creative Generation, including content creator tooling (Section~\ref{sec:creator-tooling}) and personalized narrative illustration (Section~\ref{sec:storyboard}). Beyond these generative applications, the evaluation protocol itself is independently useful: the identity element extraction pipeline provides a reusable diagnostic for user identity modeling (Section~\ref{sec:identity-modeling}), the retrieval-based metrics support emerging work on generative recommendation (Section~\ref{sec:gen-rec}), and the multi-axis evaluation suite offers a standardized testbed for personalized multimodal agents (Section~\ref{sec:agents}) and reasoning-interleaved generation more broadly (Section~\ref{sec:reasoning}). We describe each use case in turn.

\subsection{Personalized Advertising and Product Presentation}
\label{sec:advertising}

E-commerce platforms currently serve static catalog images to all users regardless of individual context. The Personalized Scene Generation task maps directly onto this setting: given a target product and a user's review history, the model must render a scene that reflects the user's inferred lifestyle and preferences. For instance, the same kitchen appliance should appear in a rustic farmhouse countertop for a user whose reviews concentrate on homesteading and organic cooking, but in a minimalist urban apartment for a user whose reviews concentrate on compact living and modern design. Practitioners can adopt the retrieval and recommendation metrics in our evaluation protocol to benchmark generative models for personalized product imagery before deployment, providing a reproducible offline proxy for engagement metrics such as click-through rate.

\subsection{Personalized Virtual Try-On and Lifestyle Staging}
\label{sec:tryon}

Virtual try-on methods typically composite a garment onto a reference body under controlled pose and lighting. Our benchmark generalizes this setting by removing the assumption of a fixed reference body and instead requiring the model to infer the staging context entirely from the user's history. For instance, the same jacket should appear on a mountain trail for an outdoor fitness enthusiast and in a city commute scene for an urban professional, as determined by their respective review histories. Unlike existing try-on benchmarks, which evaluate garment fidelity in isolation, our retrieval and identity element metrics measure whether the staged context is appropriate for the target user, a dimension that current benchmarks do not address.

\subsection{Object-to-Scene Generation as Inverse Image Editing}
\label{sec:inverse-editing}

Conventional image editing takes a nearly complete scene as input and applies a local modification (e.g., removing, replacing, or inserting an object) according to a textual instruction. Personalized Scene Generation inverts this direction: the input is an isolated object, and the task is to synthesize an entire surrounding scene from scratch, conditioned on the user's history rather than on a pre-existing canvas. This object-to-scene formulation introduces challenges absent from standard editing benchmarks, including global scene layout planning, coherent background generation, and contextual plausibility with respect to a user's inferred identity. Our benchmark thus provides a complementary evaluation axis for the editing community, measuring progress on outward scene construction around a given object rather than inward modification within a given scene.

\subsection{Content Creator Tooling}
\label{sec:creator-tooling}

Social media creators cultivate distinctive visual identities through their posting histories, and generated content that aligns with this identity is more likely to integrate naturally into their feeds. The Personalized Creative Generation task directly measures whether a model can extend a creator's established aesthetic to a novel topic, making it a natural evaluation harness for AI-assisted content creation tools. The contrastive retrieval metric, which tests whether a generated image can be attributed to its creator against stylistic confounders, provides a principled signal for model iteration. This enables tool builders to quantify style fidelity without relying on costly human evaluation.

\subsection{Personalized Storyboard and Narrative Illustration}
\label{sec:storyboard}

Generating illustrations consistent with a particular user's visual taste is central to personalized storyboarding, children's book creation, and narrative content generation. The Personalized Creative Generation task maps directly onto this setting: the topic serves as the narrative prompt, and the user history defines the illustrative style. The intra-category contrastive retrieval metric, which evaluates whether a generated image can be attributed to its user against confounders that share the same topical category but differ in visual style, provides a diagnostic for style consistency across generated illustrations, a property essential for coherent visual storytelling but difficult to quantify with existing benchmarks.

\subsection{User Identity Modeling and Understanding}
\label{sec:identity-modeling}

The identity element F1 metric introduced in our evaluation protocol extracts recurring objects, brands, settings, and stylistic cues from a user's history and measures their presence in the generated image. This metric is useful beyond image generation: researchers studying user modeling, preference elicitation, or profile summarization can adopt the identity element extraction and scoring pipeline as a standalone diagnostic for evaluating inferred user representations. Unlike aggregate embedding-based similarity measures, identity element F1 is interpretable at the per-element level, counting the concrete identity cues the model captured and the ones it missed.

\subsection{Personalized Recommendation via Generative Retrieval}
\label{sec:gen-rec}

Traditional recommendation pipelines retrieve from a fixed item catalog, but an emerging line of work explores generating candidate items rather than retrieving them. Our benchmark supports this paradigm by pairing user histories with a retrieval-based evaluation protocol that scores generated images on whether they carry sufficient user-specific signal to identify the correct user or product. This makes the benchmark useful as a testbed for generative recommendation, complementing standard rating-based or click-based evaluation with a visual fidelity axis that existing recommendation benchmarks do not provide.

\subsection{Evaluation of Personalized Multimodal Agents}
\label{sec:agents}

As multimodal agents are increasingly deployed in interactive settings where they produce visual outputs tailored to individual users, a standardized evaluation protocol for user-conditioned image generation becomes essential. Our benchmark provides such a protocol: the combination of fidelity, user alignment, and output quality axes jointly measures whether an agent's visual outputs are both high-quality and user-appropriate. Researchers building personalized assistants, avatar generators, or interactive design copilots can adopt the evaluation suite to test whether their agent genuinely adapts to user identity or merely produces generic outputs that satisfy surface-level quality constraints.

\subsection{Multimodal Reasoning and Planning}
\label{sec:reasoning}

\textsc{Pearl}'s interleaved reason-then-reflect architecture is not specific to personalized image generation. Any task that requires translating a complex, distributed multimodal context into a concrete output specification, and then verifying that specification against rendered evidence, can benefit from the same two-stage structure. Our evaluation protocol separately measures planning quality through fidelity metrics and reflection quality through ablation of the reflector stage, providing a reusable template for evaluating reasoning-interleaved generation in other settings (e.g., personalized document layout, slide deck generation, or interior design rendering).

\end{document}